\documentclass[10pt,twocolumn,letterpaper]{article}

\usepackage{cvpr}              % To produce the CAMERA-READY version
\usepackage[dvipsnames]{xcolor}

\definecolor{cvprblue}{rgb}{0.21,0.49,0.74}
\usepackage[pagebackref,breaklinks,colorlinks,citecolor=cvprblue]{hyperref}
\usepackage{multirow}
\usepackage{dutchcal}

\def\paperID{1535} % *** Enter the Paper ID here
\def\confName{CVPR}
\def\confYear{2024}

\title{Learning Inclusion Matching for Animation Paint Bucket Colorization}

\if 0
\author{
Yuekun Dai$^{1}$ $\,\,\,\,$   Shangchen Zhou$^{1}$ $\,\,\,\,$   Qinyue Li$^{1}$$\,\,\,\,$  Chongyi Li$^{2}$ $\,\,\,\,$  Chen Change Loy$^{1}$ \\
$^{1}$S-Lab, Nanyang Technological University$\,\,\,\,$ $^{2}$Nankai University \\
{\tt\small \url{https://ykdai.github.io/projects/InclusionMatching}}
}
\fi
\author{
Yuekun Dai $\,\,\,\,$   Shangchen Zhou $\,\,\,\,$   Qinyue Li $\,\,\,\,$  Chongyi Li $\,\,\,\,$  Chen Change Loy \\
S-Lab, Nanyang Technological University \\
{\tt\small \url{https://ykdai.github.io/projects/InclusionMatching}}
}

\if 0
\and
Second Author\\
\and
{\tt\small secondauthor@i2.org} \vspace{-6mm}
\fi

\newcommand{\pbc}{\textit{PaintBucket-Character}}

\begin{document}

% \if 0

\twocolumn[{%
\renewcommand\twocolumn[1][]{#1}%
\maketitle
\begin{center}
    \centering
    \captionsetup{type=figure}
    \vspace{-8mm}
    \includegraphics[width=1.0\textwidth]{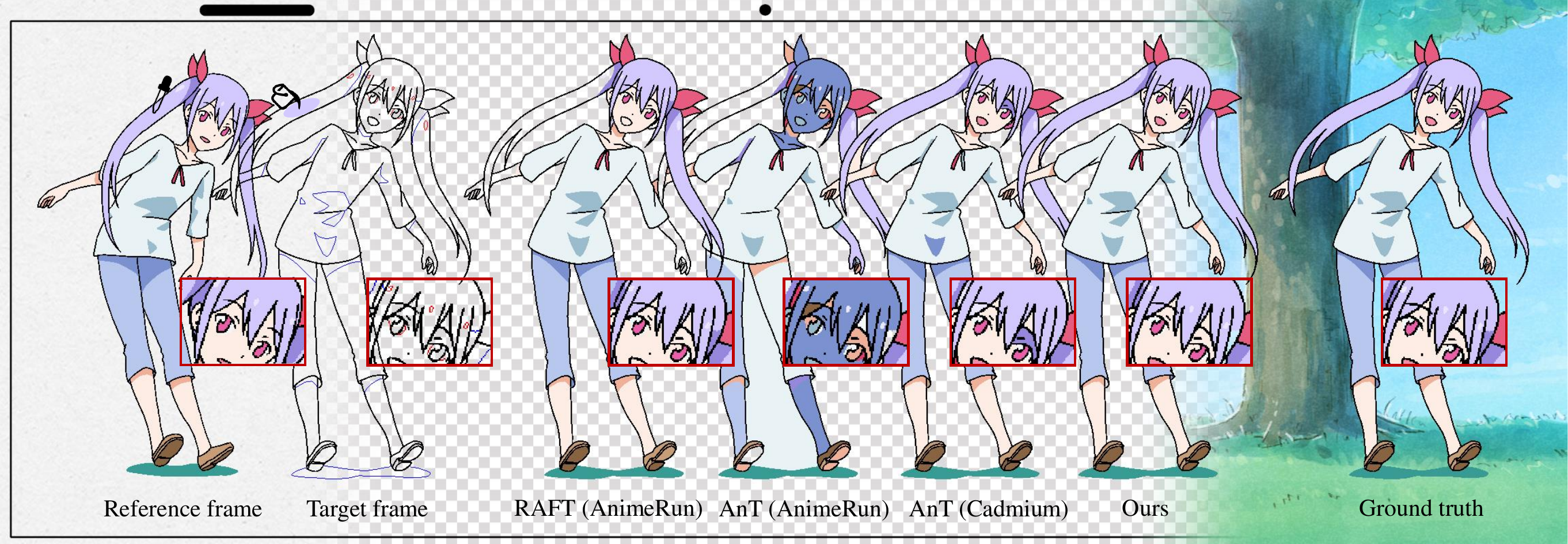}
    \vspace{-6mm}
    \captionof{figure}{In the animation industry, digital painters use paint bucket tool to colorize drawn line arts frame by frame. Our proposed pipeline streamlines this process by requiring the painters to colorize just one frame, after which the algorithm autonomously propagates the color to subsequent frames, enabling automatic colorization. Compared with optical-flow-based method RAFT~\cite{teed2020raft} and segment-matching-based method AnT~\cite{AnT}, our method can achieve more robust results on challenging cases such as one-to-many matching, large deformation, and tiny region colorization. In this figure, RAFT is trained on Sintel dataset~\cite{sintel} and finetuned on AnimeRun~\cite{siyao2022animerun}. We use the most frequent color in each segment to colorize each line-enclosed region. \copyright drawn by Nicca (Sriprachum Kongwisawamit), used with artist permission. }
    \label{fig:production_pipeline}
\end{center}%
}]
%\maketitle

\begin{abstract}

Colorizing line art is a pivotal task in the production of hand-drawn cel animation. This typically involves digital painters using a paint bucket tool to manually color each segment enclosed by lines, based on RGB values predetermined by a color designer. This frame-by-frame process is both arduous and time-intensive.
Current automated methods mainly focus on segment matching. This technique migrates colors from a reference to the target frame by aligning features within line-enclosed segments across frames. However, issues like occlusion and wrinkles in animations often disrupt these direct correspondences, leading to mismatches.
In this work, we introduce a new learning-based inclusion matching pipeline, which directs the network to comprehend the inclusion relationships between segments rather than relying solely on direct visual correspondences.
Our method features a two-stage pipeline that integrates a coarse color warping module with an inclusion matching module, enabling more nuanced and accurate colorization.
To facilitate the training of our network, we also develope a unique dataset, referred to as \pbc. This dataset includes rendered line arts alongside their colorized counterparts, featuring various 3D characters.
Extensive experiments demonstrate the effectiveness and superiority of our method over existing techniques.

\vspace{-4mm}
\end{abstract}    
\section{Introduction}
\label{sec:intro}

Colorizing line art is a critical step in hand-drawn animation production. Initially, key animators create the characters' keyframes, followed by in-between animators who generate intermediate frames and refine keyframes for seamless connection between each stroke, as described in~\cite{siyao2023inbetween}. These line arts are then scanned into binary images, featuring a primary color palette of red, blue, green, and black, which represents highlights, shadows, special instructions (\eg, eye irises), and normal lines respectively.
These scanned images, with regions of transparent background as shown in Fig.~\ref{fig:production_pipeline}, are forwarded to digital painters. The painters use a paint bucket tool and a designated color palette to fill each line-enclosed segment, including the color lines, which is known as paint bucket colorization.
Finally, special effect artists add visual enhancements, such as anti-aliasing, to blend character animations with backgrounds, culminating in the final video composition, detailed in~\cite{ae_for_animation}.

Line art colorization is a notably laborious process. For instance, even the relatively simple scenario in Fig.~\ref{fig:production_pipeline}, involving 93 segments, still requires hundreds of manual clicks. To mitigate this demanding task, several commercial software applications like Retas Studio Paintman, OpenToonz, and CLIP Studio Paint have been developed. These tools provide helpful features that allow users to easily adjust assigned colors, fill small areas, and aid in line colorization. Despite these advancements, fully automated colorization has yet to be realized.

\begin{figure}[t]
  \centering
   \includegraphics[width=1.0\linewidth]{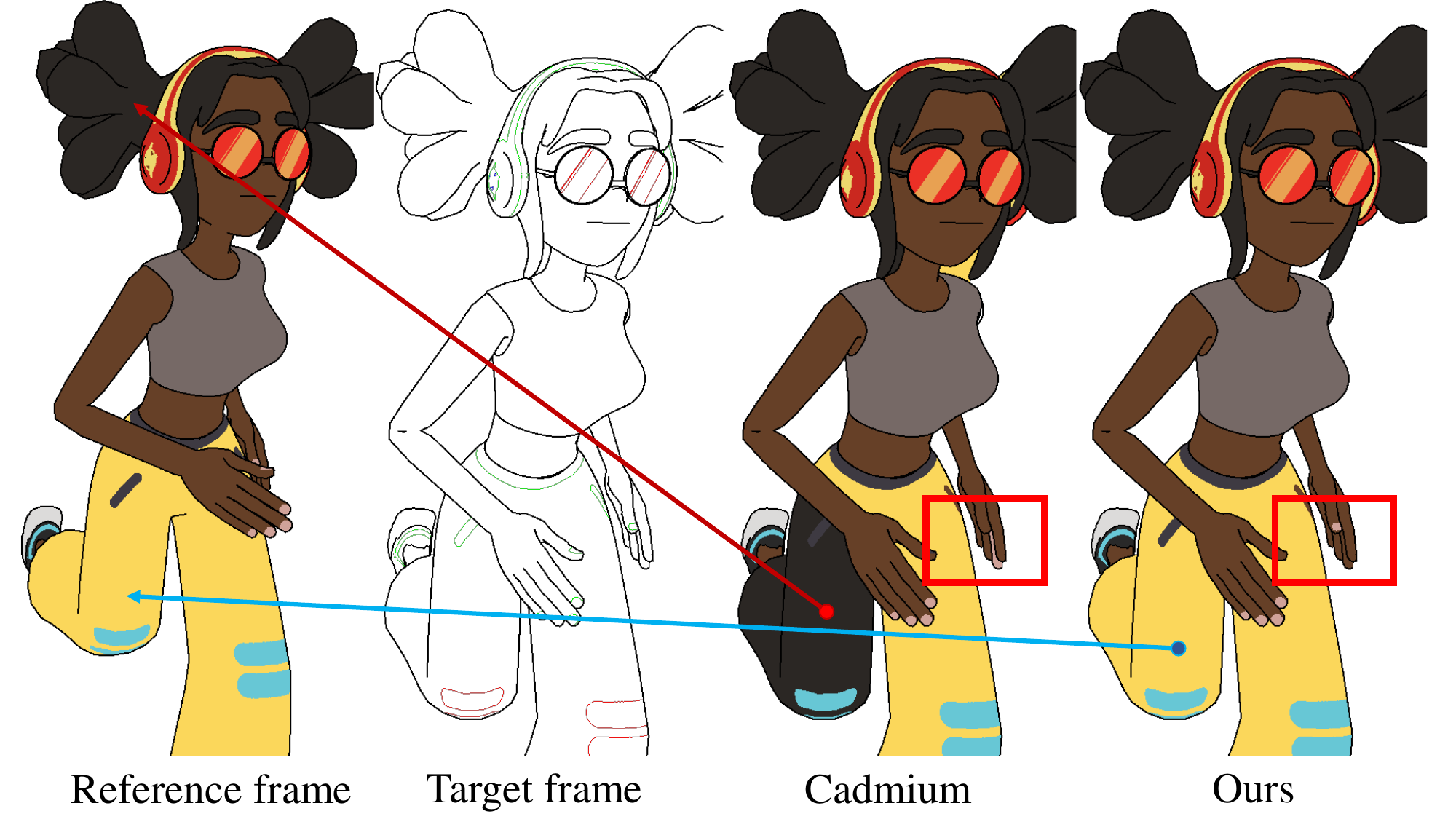}
   \vspace{-6mm}
   \caption{Methods relying on segment matching typically seek the most similar segment across frames, yet challenges arise in scenarios involving occlusion and wrinkles. This is particularly evident in tiny segments, as highlighted in the red box, where disruptions in correspondence lead to mismatches. Our innovative approach, based on inclusion matching, addresses this issue by estimating the inclusion relationship rather than pursuing direct correspondence, as illustrated by the comparison of red and blue matching lines.}
   %\lichongyi{please explain the red boxed and red and blue lines in this caption.}
   \vspace{-9mm}
   \label{fig:ant_failure_case}
\end{figure}

To address the challenges of line art colorization, numerous segment-matching methods have been developed for identifying corresponding segments in reference and target frames. Traditional graph-based approaches~\cite{maejima2019graph,liu_shape_2023,liu2020shape,zhu-2016-toontrack,zhang_excol_2012} consider each segment as a graph node and establish edges based on adjacency relationships. These methods employ optimization techniques to minimize topological similarity loss among corresponding nodes. However, they are time-intensive, often taking several minutes to process a single image with around 100 segments.
Casey~\etal~\cite{AnT} introduced the Animation Transformer (AnT) and its application Cadmium, which significantly improved segment matching. AnT standardizes the size of each segment and employs convolutional neural networks (CNNs) to extract features, transforming the image into a segment sequence. This converts segment matching into a sequence-to-sequence problem.
Although AnT excels in aligning segments between adjacent frames featuring minimal motion, it struggles in more complex scenarios, particularly those with occlusion or pronounced motion. During occlusions, interlaced lines tend to fragment segments into multiple pieces, as depicted in Fig.~\ref{fig:ant_failure_case}. This fragmentation disrupts the strict correspondence between segments, rendering the task of segment matching inherently problematic and poorly defined.

\begin{figure*}[h]
  \centering
    \vspace{-2mm}
   \includegraphics[width=0.9\linewidth]{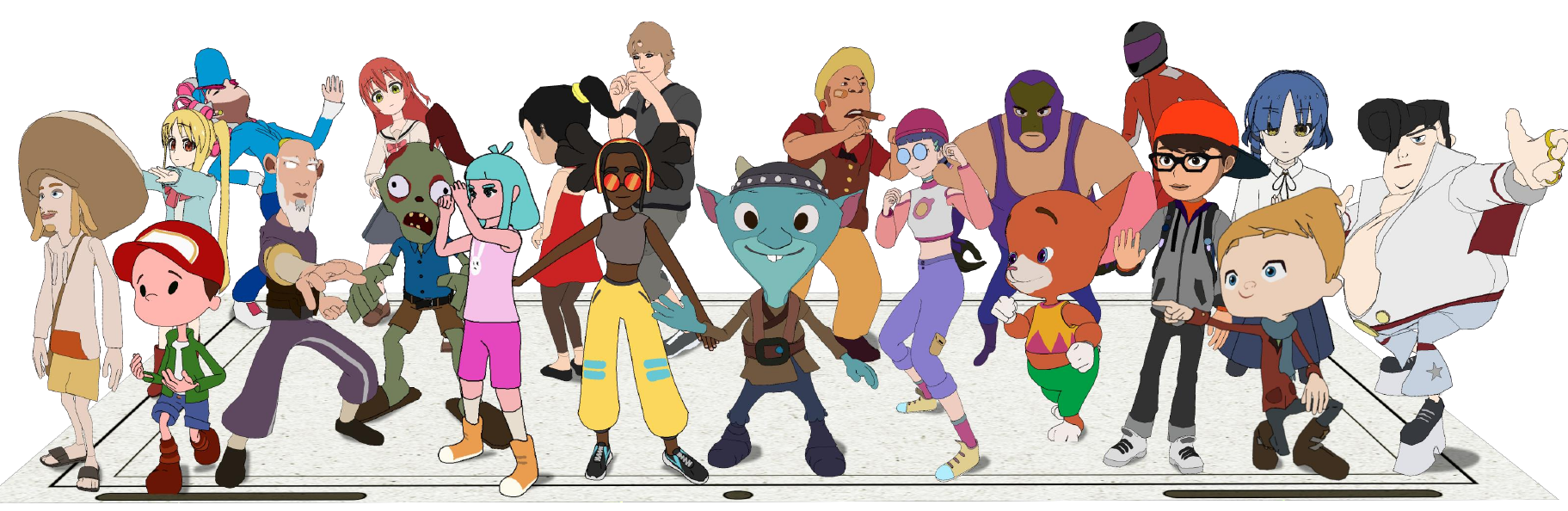}
   \vspace{-3mm}
   \caption{Several examples of our rendered characters in \pbc~dataset.}
   \vspace{-7mm}
   \label{fig:dataset}
\end{figure*}

In this study, we introduce the novel notion of `\textbf{inclusion matching}', which obviates the necessity for exact segment-to-segment correspondence. Instead of direct segment matching, our approach computes the likelihood of each segment in a target frame being included within a specific region of a reference frame.
Our methodology is unique in that it follows a \textbf{coarse-to-fine colorization} pipeline, inspired by the approach of skilled digital painters. This approach typically begins with coloring larger, more noticeable segments across all frames before filling in details of smaller segments in each frame.
Motivated by this workflow, our pipeline first computes optical flow between line sketches, warping colors from a reference to a target frame to achieve a preliminary, coarse colorization. 
This is followed by the application of an inclusion matching module, which refines the colorization result of tiny segments.

The sparse nature of line art requires special formulation in our approach.
Specifically, recognizing the necessity for a large receptive field in extracting line art features, we employ a U-Net structure~\cite{unet} to process concatenated color and line features. 
To address the difficulty in matching segments under large deformations, we extract line semantic features using CLIP~\cite{clip} and concatenate them with features from the U-Net's bottleneck. This allows the network to match the region with the same semantic information across frames.
Finally, we introduce super-pixel pooling at each segment to transfer the image into tokens for sequence-to-sequence matching. 

Dataset plays a key role in the success of our method. To train and evaluate our method, we select 22 character models from Mixamo~\cite {mixamo} and Aplaybox~\cite{aplaybox}. 
We mimic real intermediate line art data on the animation coloring production line. We disable the anti-aliasing and provide the color lines which is quite common in animation production.
Then we use 3D software Cinema 4D to synthesize a specially-designed cartoon character colorization dataset named \pbc. 

The contributions of our work include the introduction of the \pbc~dataset tailored for paint bucket colorization tasks, the novel inclusion matching pipeline addressing the challenge of lacking strict segment correspondence, and a new coarse-to-fine baseline that outperforms existing solutions, particularly in scenes with occlusion. Our experiments demonstrate that our method surpasses the performance of recent AnT application Cadmium, and outperforms the results presented by Li~\etal.~\cite{siyao2022animerun}. Notably, the advantages of our approach extend beyond dataset improvements, highlighting the effectiveness of our newly designed inclusion matching pipeline and the coarse-to-fine baseline in the context of animation colorization.

\section{Related Work}
\label{sec:formatting}

%----------------------------------------------------------------------
% \subsection{Scattering flare removal}
\vspace{2mm}
\noindent{\bf Line Art Colorization.}
% Reference-based line-art colorization
% scribble-based line-art colorization
% Animation colorization
Line art colorization aims to create reasonable colors in the sketch's blank regions. To achieve precise control, many user guidance types are designed including text~\cite{kim_tag2pix_2019,zouSA2019sketchcolorization}, scribble~\cite{sykora2009lazybrush, xia-2018-invertible, Filling2021zhang, cao2021line}, colorized reference~\cite{sykora2004unsupervised,sykora2009rigid,chen2020active, lee2020reference,wu2023self,zhang2021line,controlnet,akita2023hand,cao2023animediffusion}. 
These methods mainly focus on pixel-wise estimation and cannot fill flat accurate RGB values assigned by the color designer in line arts' segments, which always leads to flickering and color bleeding. 
Besides, in the post-processing, the special effect artists often need to select specific colors to create various effects such as ramp or glow for the character's specific region. Thus, current line art colorization methods are still difficult to be merged into animation industrial processes.

\vspace{2mm}
\noindent{\bf Animation-related Dataset.} To facilitate hand-drawn animation production, many real and rendered datasets are proposed. 
To boost video interpolation performance, Li~\etal~\cite{siyao_deep_2021} propose the ATD-12K dataset which consists of triplet frames from 30 real animation movies. 
Due to the absence of correspondence labels for real animation, addressing this challenge has led to the adoption of 3D rendering techniques.  MPI-Sintel~\cite{sintel} and CreativeFlow+~\cite{shugrina2019creative} provide different annotations such as optical flow, segmentation labels, and depth maps for multiple applications using Blender. 
To overcome the domain gap between rendered images and real 2D cartoons, Li~\etal~\cite{siyao2022animerun} developed the AnimeRun dataset using open-source 3D animations and their rendered optical flow. 
It improves segment matching accuracy and handles complex scenes. 
%ATD-12K, Sintel, CreativeFlow+, AnimeRun, MixamoLine240

\vspace{2mm}
\noindent{\bf Segment-based Animation Colorization.}
%animation colorization + segment matching
In the cleaning-up process of the animation industry, animators enclose each line manually, making segments serve as basic units for colorization. Thus, how to calculate segment-level semantic information and correspondence becomes the key in animation colorization. Traditional methods~\cite{maejima2019graph,liu_shape_2023,liu2020shape,zhu-2016-toontrack,zhang_excol_2012} regards segment as node and adjacency relationship as edge which transfer the segment matching to the graph optimization problem. 
% Dang -> Hu moments -> UNet -> superpixel pooling
Recently, Dang~\etal~\cite{dang_correspondence_2020} use Hu moments~\cite{hu1962visual} to extract each segment's feature to replace the RGB value and apply a UNet structure~\cite{unet} to obtain the feature map. 
Subsequently, the feature is averaged in each segment to calculate the distance across frames for matching. 
% Casey -> use CNN to extract segment feature add transformer 
Referring to the success of Transformer~\cite{vaswani2017attention} in image matching~\cite{sarlin20superglue}, Casey~\etal~\cite{AnT} apply multiplex transformer to aggregate information across frames to obtain more accurate results.
Besides, Casey~\etal replace the Hu moments algorithm with a segment feature extraction CNN which crop and resize each segment to fixed resolution first.
% AnimeRun
Since both Dang~\etal and Casey~\etal's methods and dataset are not publicly available, AnimeRun~\cite{siyao2022animerun} remains the sole benchmark for segment matching available to researchers.

\begin{table*}[h]
\label{tab:data_compare} 
%\begin{center}
\centering
\vspace{-3mm}
\caption{A comparison between our \pbc~dataset and previous segment matching datasets. Since AnT~\cite{AnT}'s dataset is not publicly available, we use `-' to mark the unknown information. AnT does possess a private hand-drawn dataset with 3578 frames; however, it does not provide train-test-split information, indicated by `+'. Compared with previous open-source datasets, our dataset is more diverse and provides color lines that are closer to real hand-drawn animation. No anti-aliasing provides extra semantic information for each segment, which enables the network to learn better semantic information rather than segment shape similarity.}
\vspace{-3mm}
\resizebox{0.95\textwidth}{!}{
\begin{tabular}{lccccccccc}

\toprule
\multicolumn{1}{l}{ \multirow{2}*{Dataset} }& \multicolumn{5}{c}{Statistics} &\multicolumn{4}{c}{Annotations}\\
%\cline{2-8}
\cmidrule(r){2-6}
\cmidrule(r){7-10}
\multicolumn{1}{l}{}&  Train & Test & Clips & Seg per Fr. & No anti-aliasing & Optical flow & Index label & Color label  & Color line\\
\toprule
AnimeRun~\cite{siyao2022animerun}             &  1,760& 1,059& 20+10& 237    &   $\times$&$\checkmark$&$\times$&$\times$&$\times$\\
AnT~\cite{AnT}~(private)  &  9,900+& 1,100+& -& $<$ 50      &   -&$\times$&$\checkmark$&$\checkmark$&$\times$\\
Ours                   &  11,345& 3,200& 180+170& 169     &   $\checkmark$&$\times$&$\checkmark$&$\checkmark$&$\checkmark$\\
\hline
\end{tabular}
}
\vspace{-5mm}
%\end{center}
\end{table*}

\section{Paint Bucket Colorization Dataset}
AnimeRun~\cite{siyao2022animerun} employs rendered optical flow to calculate segment correspondence, leading to frequent one-to-many and many-to-one matching in the dataset.
This results in inaccurate benchmark evaluations, as each segment may have multiple potential ground-truth correspondences.
Furthermore, the limited dataset provided by AnimeRun, consisting of only 1,760 training images, increases the risk of network overfitting.
Meanwhile, due to the anti-aliasing processing of AnimeRun's rendered images, accurate color labels cannot be obtained in each segment. Consequently, AnimeRun faces difficulties in serving as a paint bucket colorization dataset directly.

To address these challenges, we introduce a new dataset~\pbc, comprising 11,345 training images and 3,200 test images. The test set includes 3,000 3D rendered frames and 200 hand-drawn frames.
Notably, our dataset focuses solely on character animations, with consideration for the common practice among animators of just drawing foreground character animations while employing 3D models or separate paintings for the background.
Our dataset focuses on character animations, considering the industry practice where animators typically sketch only the foreground character movements, while backgrounds are created using separate 3D models or paintings.
Different from AnimeRun, our evaluation focuses on segments' color accuracy rather than traditional matching accuracy.
To build our dataset, we sample 22 characters from Mixamo~\cite{mixamo} and Aplaybox~\cite{aplaybox} and set 12 characters for training and 10 for testing.
Then, we replot the UV map for these character models in flat color style and use Cinema 4D's \textit{Sketch and Toon Effect} to produce the line art.
To ensure consistent color in each segment, we disable the anti-aliasing and change the texture sampling method to the nearest.
As depicted in Fig.~\ref{fig:dataset}, our dataset showcases a diverse array of characters encompassing both Japanese and Western cartoon styles.

\begin{figure}[t]
  \centering
   \includegraphics[width=0.8\linewidth]{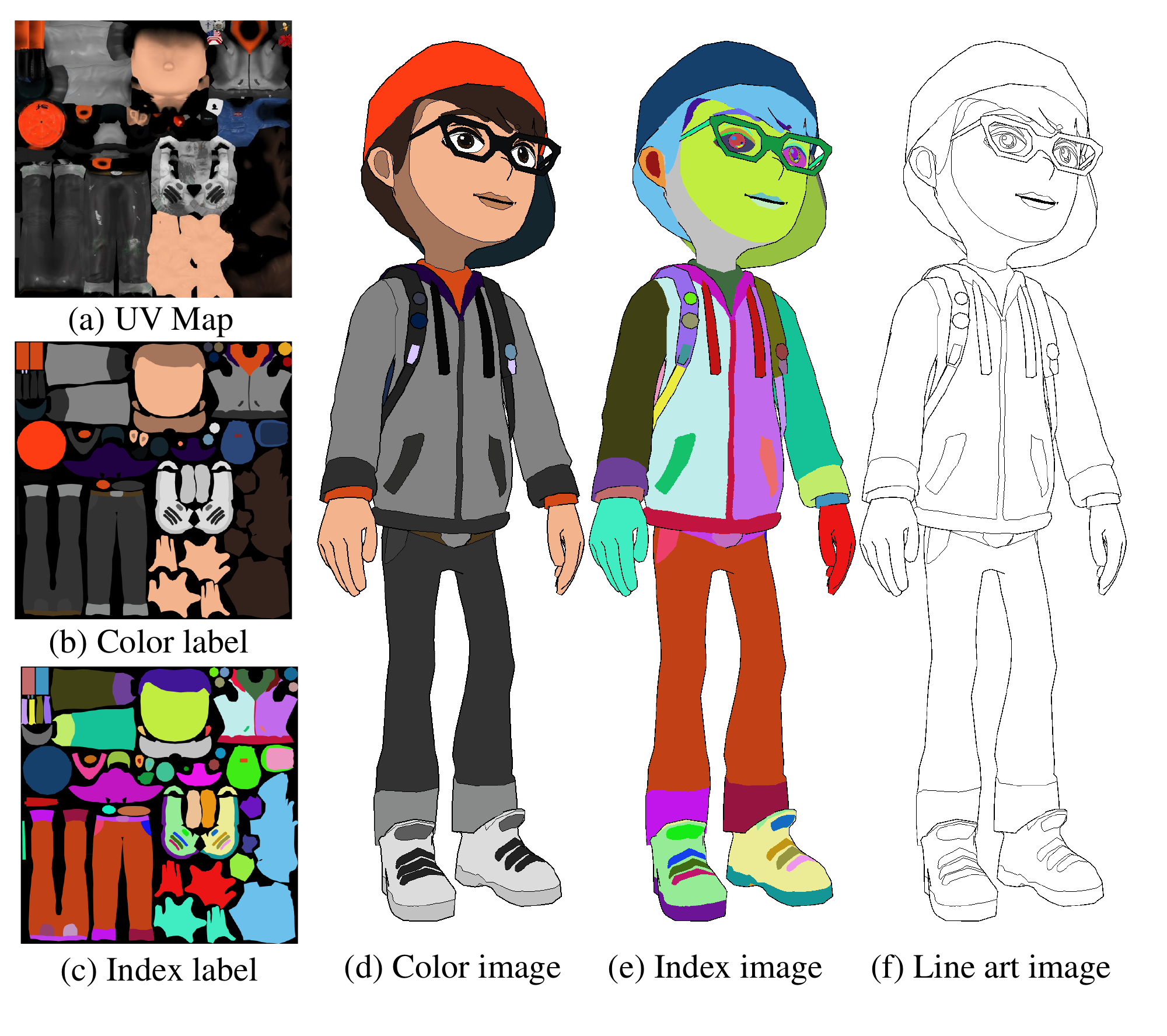}
   \vspace{-4mm}
   \caption{Overview of our synthetic data generation pipeline. We extract the UV map (a) and the used UV region from the character. Then, we use a paint bucket tool to fill the used UV region, creating color label (b) and index label (c). Label images (b) and (c) are then pasted back to the 3D meshes to create flat color characters (d) and (e), respectively. Finally, we post-process the index image (e) to obtain the rendered line art image (f).}
   \vspace{-6mm}
   \label{fig:data_generation_pipeline}
\end{figure}

\subsection{Synthetic Dataset Construction}
% 11345 train  train res: 1920  test: 1024
% 3000+200 test
To include temporal and occlusion relationships in our dataset, we apply Mixamo's actions to our characters. For each character, we add five types of actions and animate the camera motion to capture different viewpoints of the character in each sequence. We classify the camera positions into three types: face shot, long shot, and close-up, in which we separately focus on the face, whole body, and other parts of the body. Following this pipeline, we obtain 15 clips for each character and form 180 clips for training and 150 clips for test. In the animation industry, keyframes always have a low frame rate~\cite{siyao_deep_2021}. Thus, in each clip, we set the frame rate to 10 to mimic real animation and bring larger motion and deformation. Meanwhile, to address the fine scale of character details such as pupils and buttons, we use 1920$\times$1920 as our training size and 1024$\times$1024 as our test set size to provide clear segments in these areas.

\noindent{\bf Flat Color Style Rendering.}
To render characters in a flat color style, we use Cinema 4D to extract a UV mesh map and set unused regions as black. Then, we draw contour lines of character details such as pockets and shoelace on the UV mesh. Based on this original UV map, we use a paint bucket tool to colorize the mesh as the color label. To obtain a more precise matching relationship for training, we connect the label with the same semantic information (such as palm and back of the hand) and use the connected components algorithm for each color to extract the index label as shown in Fig.~\ref{fig:data_generation_pipeline}. Since we aim to achieve accurate line art to separate each segment, we calculate the difference between each color's segment and its erosion as the black lines. This operation ensures that each line-enclosed segment contains clear semantic information and comprises only one color.

\begin{figure}[t]
  \centering
   \includegraphics[width=1.0\linewidth]{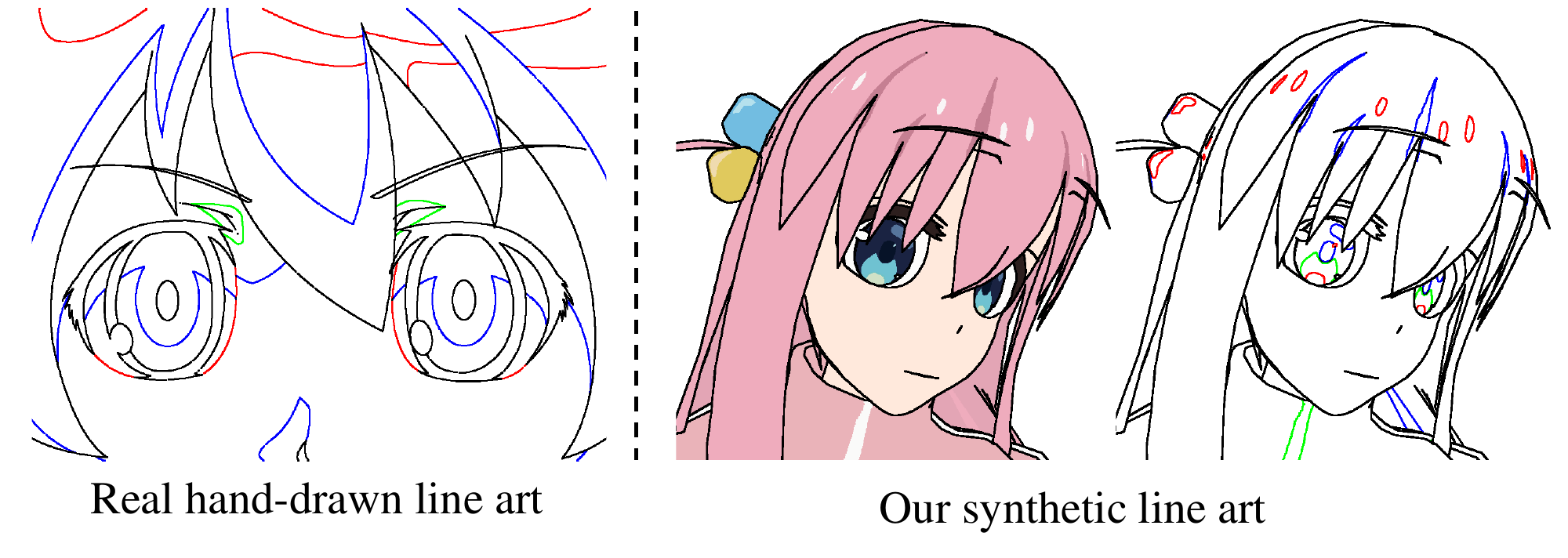}
   \vspace{-8mm}
   \caption{Our simulation method can generate color lines that closely mimic hand-drawn line arts. Red, blue, and green 
 lines represent highlights, shadows, and other instructions respectively.}
   \vspace{-7mm}
   \label{fig:line_generation_pipeline}
\end{figure}

\noindent{\bf Color Line Extraction.}
To simulate color lines which are common in real hand-drawn animation's shadow and highlight regions, we mark these highlight and shadow colors and erode these regions to create color lines, as shown in Fig.~\ref{fig:line_generation_pipeline}.
Compared with AnimeRun~\cite{siyao2022animerun}, our dataset features more realistic line art and provides clear semantic information for each segment. Since our characters are rendered with discrete colors, we can easily obtain specific meanings for each segment.

\subsection{Hand-drawn Dataset Collection}

Obtaining intermediate unprocessed data poses a challenge, as common commercial animations undergo post-processing techniques including anti-aliasing.
To address this challenge, we invite some professional animators to draw line art animation clips with different characters. Then, we use the paint bucket tool to colorize these line arts as our ground truth to synthesize paired line art and colorized data. Besides, animation creation software such as Retas Studio Paintman and CLIP Studio Paint all provide animation production tutorials. We collect animations in these tutorials and use the color line extraction method to synthesize the paired data. Finally, we collect a hand-drawn test dataset with 200 frames and 20 clips for evaluation.

\section{Method}
\vspace{2mm}
\subsection{\bf Inclusion Matching}  Given two line art images $L_i,L_j \in \mathbb{R}^{H \times W \times 3}$, lines separate the blank area into $N_i$ and $N_j$'s line-enclosed segments. In the reference frame $L_i$, each segment is colorized with $M$ colors whose color labels can be represented as one-hot representation $c_i \in \mathbb{R}^{N_i \times M}$. To estimate the $L_j$'s color label $\hat c_j \in \mathbb{R}^{N_j \times M}$, previous segment-matching-based methods~\cite{AnT,siyao2022animerun} mainly focus on calculating segments' visual correspondence with the matching $\textbf{m}^{j \to i} \in \mathbb{R}^{N_j \times N_i}$. 
Nevertheless, the absence of strict correspondence among segments imposes limitations on the model's learning capacity.

\begin{figure}[t]
  \centering
   \includegraphics[width=0.8\linewidth]{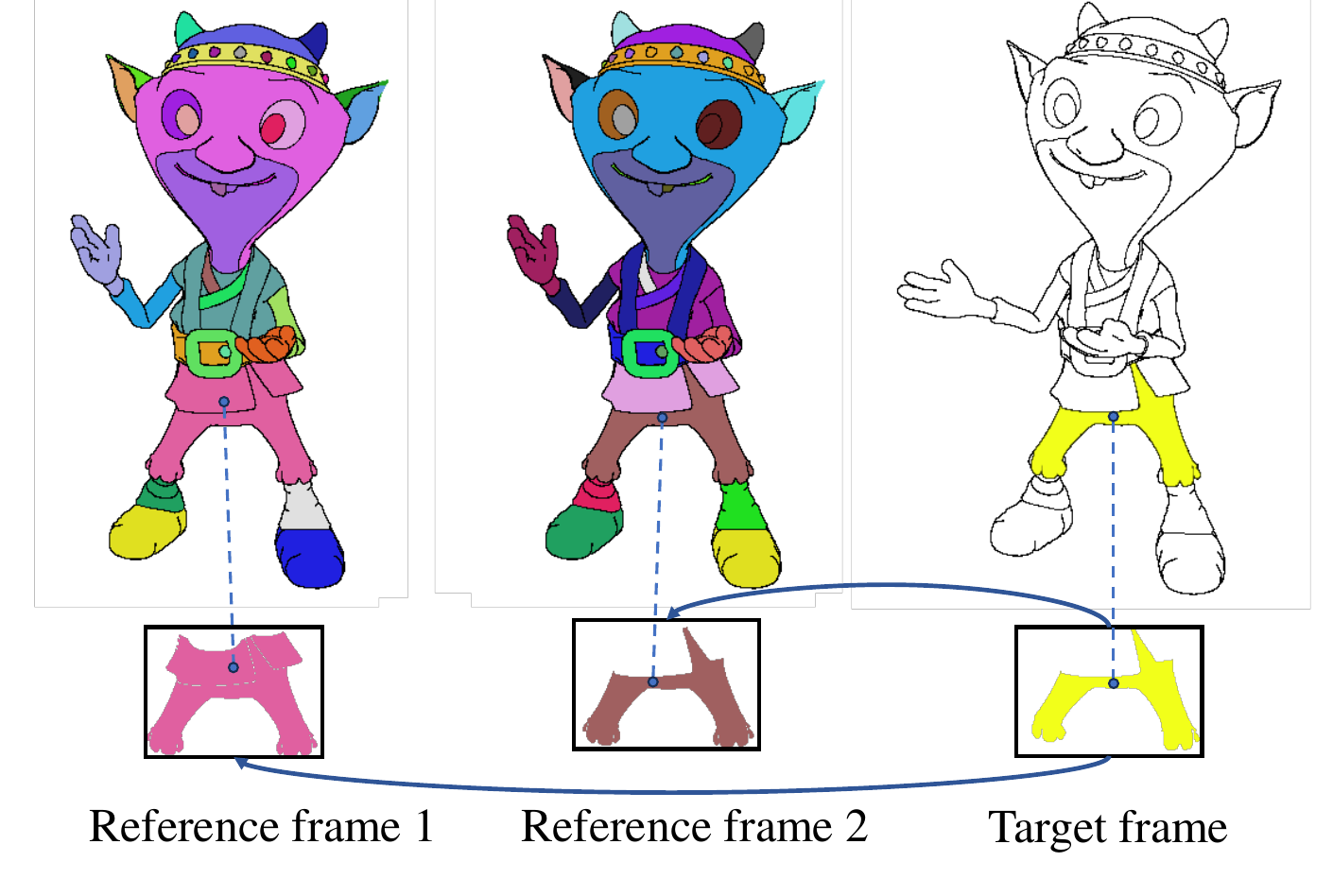}
   \vspace{-4mm}
   \caption{Our inclusion matching training pipeline employs index labels to partition the reference frame into $K_i$ regions, which are then randomly merged with adjacent index labels with a probability of 20\%. Subsequently, the model estimates the affiliation of each line-enclosed segment in the target frame with a specific region in the reference, thereby facilitating the learning of the inclusion relationship.}
   \vspace{-6mm}
   \label{fig:inclusion_matching}
\end{figure}

To tackle this challenge, we introduce a novel method called inclusion matching. In the reference frame, segments with identical semantic information are merged, creating $K_i$ large regions with different labels. As each line-enclosed segment in the target frame can only possess one color with a specific meaning, it can be identified as a subset of one of the $K_i$ regions.
Specifically, in the training process, the index label stated in Fig.~\ref{fig:data_generation_pipeline} separates the reference frame into $K_i$ region. We train a network to estimate each segment's index label $\mathbb{R}^{N_j \times K_i}$.
To enrich the diversity, we randomly merge the adjacent labels in the reference frame to create larger regions to guide the network to learn a more robust inclusion relationship, as shown in Fig.~\ref{fig:inclusion_matching}.

\subsection{\bf PaintBucket Colorization Pipeline} 

\begin{figure*}[h]
  \centering
   \includegraphics[width=1.0\linewidth]{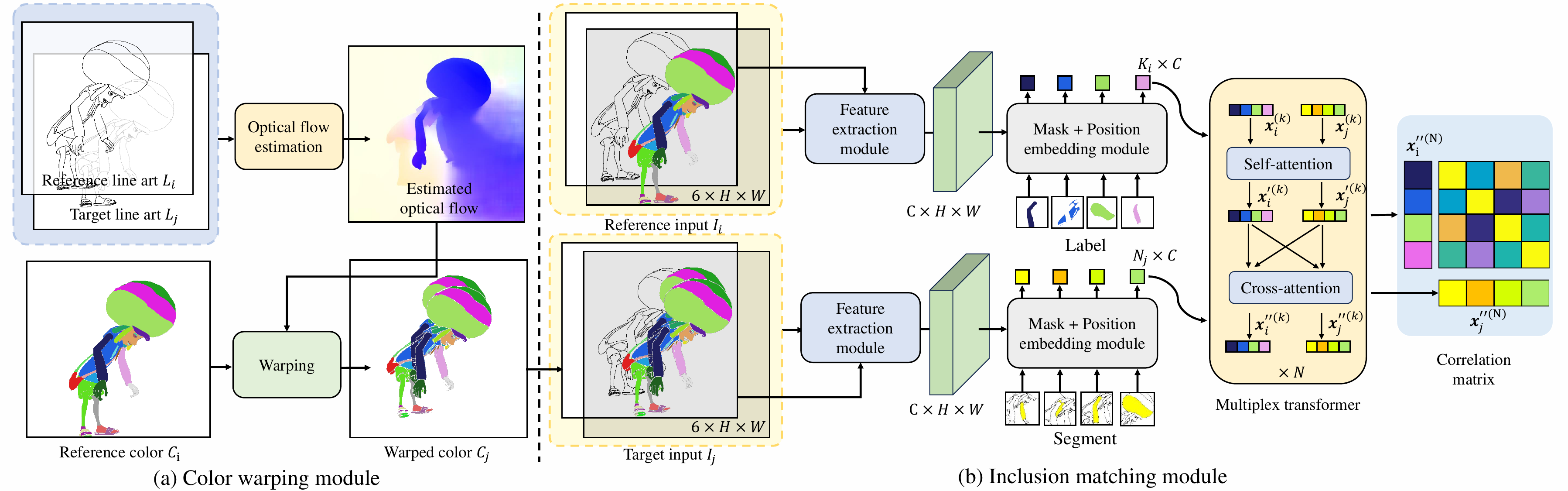}
   \vspace{-6mm}
   \caption{Our paint bucket colorization architecture. Initially, we estimate the optical flow based on the line arts and warp the reference color as the coarse colorization result. The mask and position embedding module tokenizes the segments and features extracted by the feature extraction module into a sequence, which is then fed into the multiplex transformer for information aggregation. Finally, we compute a similarity matrix between reference and target tokens to calculate matching loss and predict the final color.}
   \vspace{-6mm}
   \label{fig:pipeline}
\end{figure*}
\vspace{2mm}
\noindent{\bf Color Warping Module.}
Colorized line art carries substantial information about how segments are grouped, making it crucial for inclusion matching.
However, previous segment matching methods~\cite{AnT,siyao2022animerun} exclusively leverage the line art's features for matching and do not incorporate the colorized reference. 
% Color redistribution
To encode the color feature, we recolorize the index label image. Specifically, we partition the RGB space into $K_i$ cubes, each with a side length of $255\times K_i^{-\frac{1}{3}}$. Subsequently, we select the RGB value at the center of each cube and randomly assign them to index labels as the reference color image $C_i$.
% Color warping
We employ the optical flow estimation model RAFT~\cite{teed2020raft}, fine-tuned on AnimeRun~\cite{siyao2022animerun}, to estimate the optical flow between the reference line art $L_i$ and the target line art $L_j$. Utilizing this optical flow, we warp the reference color image $C_i$ to generate the coarse color estimation $C_j$ for the target frame.

\begin{figure}[t]
  \centering
   \includegraphics[width=1.0\linewidth]{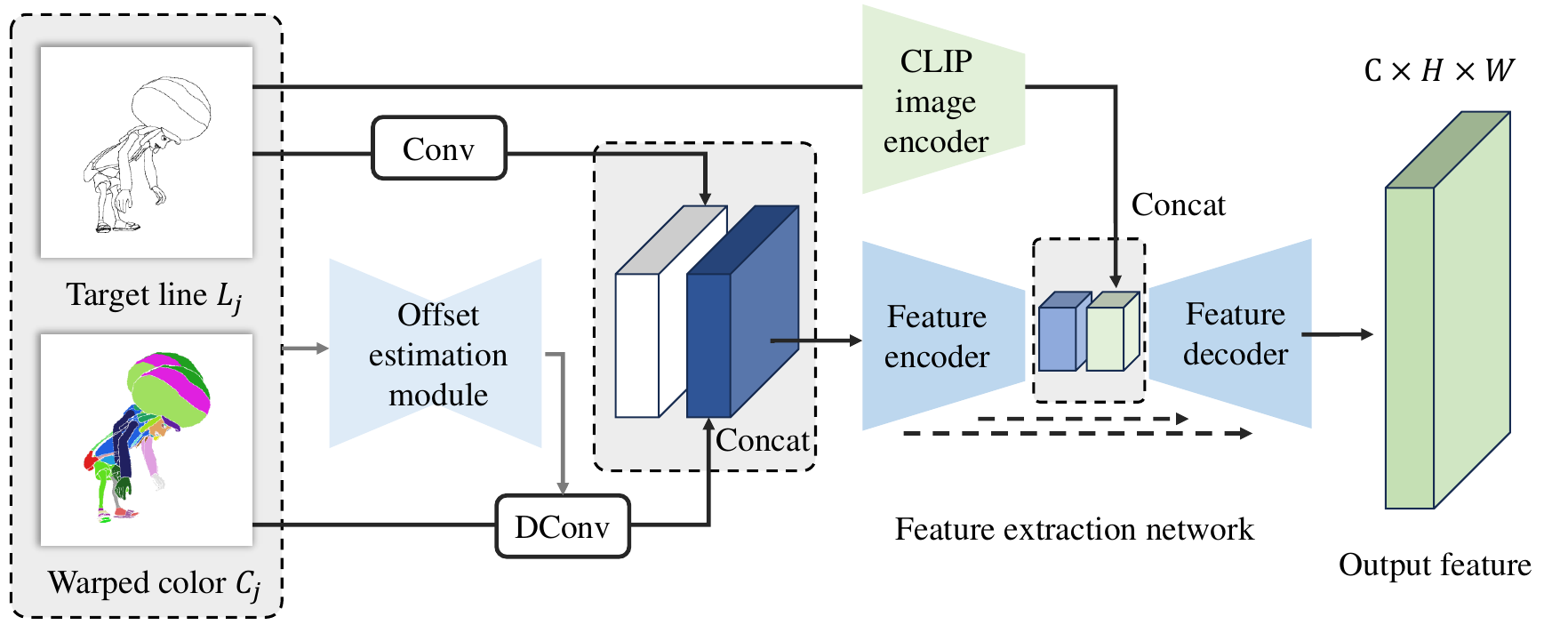}
   \vspace{-7mm}
   \caption{Overall pipeline of our feature extraction module. Deformable convolution is applied to align the color features of the target frame to the line image. Then, the line features and aligned color features are concatenated and fed to the feature extraction network in a U-Net structure. Line art's CLIP features are also concatenated to the bottleneck of the feature extraction network.}
   \vspace{-7mm}
   \label{fig:feature_extraction_network}
\end{figure}

\noindent{\bf Feature Extraction Module.}
%Since are not aligned
Lacking textual information and large deformation in animation make warped color image $C_j$ hard to precisely align with the line art in the target frame. To address this, we employ deformable convolution~\cite{dai2017deformable} on the warped color $C_j$ to extract aligned features. For a given deformable convolution kernel of $K$ sampling positions, we use $\omega_k$ and $\mathbf{p}_k$ to represent the weight and pre-specified offset. The aligned features $F_{a}$ at each position $\mathbf{p}_0$ can be expressed as:
\begin{equation}
    F_{a}(\mathbf{p}_0)=\sum_{k=1}^{K} \omega_k \cdot F_a(\mathbf{p}_0+\mathbf{p}_k+\Delta \mathbf{p}_k), 
\end{equation}
where learnable offset $\Delta \mathbf{p}_k$ can be predicted from the offset estimation network as depicted in Fig.~\ref{fig:feature_extraction_network}. 
This network, designed as a lightweight U-Net~\cite{unet} denoted as $\Phi$, is employed to expand the receptive field:
\begin{equation}
    \Delta \mathbf{p}_k =\Phi([L_j,C_j]).
\end{equation}
For the reference frame, we deactivate the offset estimation module and set the offset to zero. Subsequently, the aligned color feature is concatenated with the line feature and passed through a feature encoder network.
To guide the network in learning semantic information matching, we resize the line art image to $320 \times 320$ and utilize CLIP~\cite{clip}'s image encoder to extract features.
To preserve the feature's positional information, we choose to use the ConvNext-Large model from OpenCLIP~\cite{cherti2023reproducible}.
We only selected the initial layers of CLIP, which downsample the feature to a resolution of $40 \times 40$.
The CLIP feature is interpolated to concatenate with the encoded features.
Finally, the feature decoder decodes these concatenated features to the input's resolution with $C$ channels. 
To preserve features in high resolution, shortcuts are added between the network's encoder and decoder, following the structure of the U-Net.

\begin{figure*}[t]
  \centering
  \vspace{-3mm}
   \includegraphics[width=0.9\linewidth]{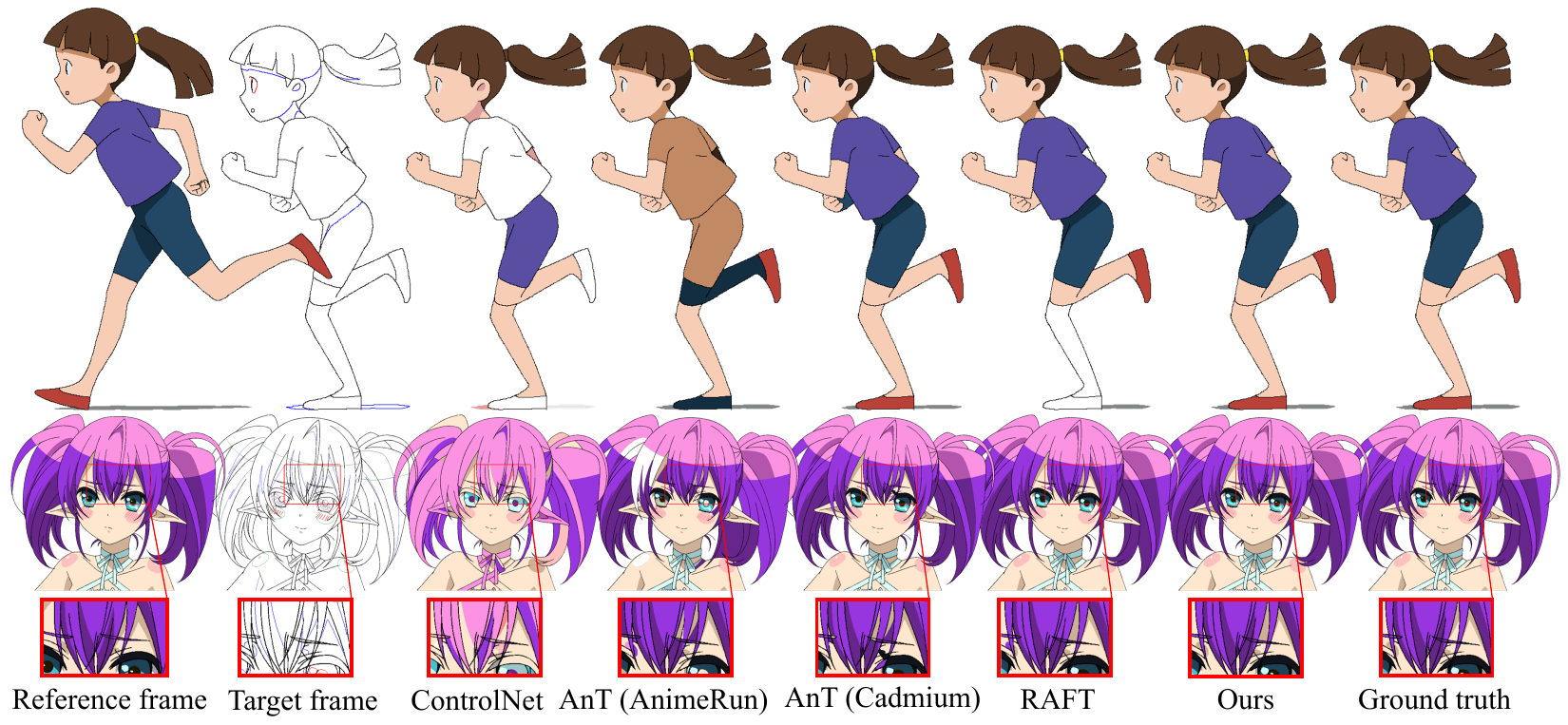}
   \vspace{-5mm}
   \caption{The visualization results of our proposed methods and other approaches on real hand-drawn animation. Compared with previous methods, our proposed approach demonstrates enhanced robustness in handling occlusion situations (\eg, hand of the running girl and shadow near eyebrows of the girl in the second row.)}
   \vspace{-3mm}
   \label{fig:real_anime}
\end{figure*}

\noindent{\bf Mask + Position Embedding Module.}
To transform different segments into tokens, we employ super-pixel pooling to average the features within each segment's mask, resulting in a $C$-dimensional visual token, denoted as $\mathbf{d}_i$.
Subsequently, a Multilayer Perceptron (MLP) layer is utilized to encode the segment's bounding box's four-dimensional coordinates, denoted as $\mathbf{\mathcal{p}}_i$, into a $C$-dimensional positional embedding.
The token $\mathbf{x}_i$ for matching is obtained by adding the positional embedding and visual token:
\begin{equation}
\mathbf{x}_i = \mathbf{d}_i + \textbf{MLP}(\mathbf{\mathcal{p}}_i).
\end{equation}
Differing from previous segment matching pipelines~\cite{AnT,siyao2022animerun}, our approach matches each line-enclosed segment in the target frame with a specific index label in the reference frame, rather than employing label-to-label matching.

\noindent{\bf Multiplex Transformer.}
Building upon \cite{sarlin20superglue,AnT}, we employ multiplex self- and cross-attention layers to aggregate tokenized features. In a self-attention layer, queries, keys, and values are derived from a single source feature:
\begin{equation}
\textbf{self-attention}(\mathbf{x}_i) = \text{softmax}\left(\frac{\mathbf{Q}_i\mathbf{K}_i}{\sqrt{D}} \right)\mathbf{V}_i,
\end{equation}
where $\mathbf{Q}_i$, $\mathbf{K}_i$, and $\mathbf{V}_i$ represent $\mathbf{x}_i$ processed by MLPs for query, key, and value. In contrast, the cross-attention layer computes keys and values from another feature:
\begin{equation}
\textbf{cross-attention}(\mathbf{x'}_i,\mathbf{x'}_j) = \text{softmax}\left(\frac{\mathbf{Q'}_i\mathbf{K'}_j}{\sqrt{D}} \right)\mathbf{V'}_j.
\end{equation}
Following each self- or cross-attention layer, the output is added to the original $\mathbf{x}_i$ or $\mathbf{x'}_i$, and then processed with a feed-forward MLP. After repeating these operations $N$ times, as illustrated in Fig.~\ref{fig:pipeline}, we obtain the final aggregated features $\mathbf{\hat x}_i\in \mathbb{R}^{K_i\times C}$ and $\mathbf{\hat x}_j\in \mathbb{R}^{N_j\times C}$.

\begin{table*}[th]
\vspace{-1mm}
\centering
\caption{
Quantitative comparison of our method with different methods. `Acc' and `Acc-Thres' denote segment-wise accuracy, providing insights into the potential workload reduction for digital painters. In `Acc-Thres,' segments smaller than 10 pixels are filtered out. `Pix-Acc,' `Pix-F-Acc,' and `Pix-B-MIoU' represent pixel-wise accuracy, foreground pixel-wise accuracy, and pixel-wise background MIoU, respectively, reflecting the visualization performance. All RAFT models undergo training on MPI-Sintel~\cite{sintel}, and `F/C' indicates the models fine-tuned on AnimeRun's colorized frames and contours.
} \label{tab:baseline}
\vspace{-3mm}
\resizebox{1.0\textwidth}{!}{
\begin{tabular}{llcccccccccccc}
\hline
                                        \multirow{2}{*}{Type} &         \multirow{2}{*}{Method}                &  & \multicolumn{5}{c}{3D rendered test set}                                                                                                           & \multicolumn{1}{l}{} & \multicolumn{5}{c}{Hand-drawn test set}                                                                                                           \\ \cline{4-8} \cline{10-14} 
                                        &                         &  & \multicolumn{1}{c}{Acc} & \multicolumn{1}{l}{Acc-Thres} & \multicolumn{1}{l}{Pix-Acc} & \multicolumn{1}{l}{Pix-F-Acc} & \multicolumn{1}{l}{Pix-B-MIoU} & \multicolumn{1}{l}{} & \multicolumn{1}{c}{Acc} & \multicolumn{1}{l}{Acc-Thres} & \multicolumn{1}{l}{Pix-Acc} & \multicolumn{1}{l}{Pix-F-Acc} & \multicolumn{1}{l}{Pix-B-MIoU} \\ \cline{1-8} \cline{10-14} 
Reference-based       & ControlNet              &  & 0.1709                        & 0.1856                        & 0.4274                        & 0.2653                        & 0.4600                         &                      & 0.1715                        & 0.1812                        & 0.2449                        & 0.2298                        & 0.2260                         \\
%                                        & IP-Adapter                  &  & \multicolumn{1}{l}{}    & \multicolumn{1}{l}{}           & \multicolumn{1}{l}{}         & \multicolumn{1}{l}{}          & \multicolumn{1}{l}{}          & \multicolumn{1}{l}{} & \multicolumn{1}{l}{}    & \multicolumn{1}{l}{}           & \multicolumn{1}{l}{}         & \multicolumn{1}{l}{}          & \multicolumn{1}{l}{}          
\cline{1-2} \cline{4-8} \cline{10-14} 
\multirow{2}{*}{Segment matching} & AnT(AnimeRun)        &  & 0.5584                  & 0.5930                         & 0.8734                       & 0.6537                        & 0.8954                        &                      & 0.6779                  & 0.7043                         & 0.9193                       & 0.7377                        & 0.9429                        \\
                                        & AnT(Cadmium)                 &  & 0.6634                  & 0.7713                         & 0.9765                       & 0.9471                        & 0.9801                        &                      & 0.7224                  & 0.7879                         & 0.9792                       & 0.9303                        & 0.9899                        \\ \cline{1-2} \cline{4-8} \cline{10-14} 
\multirow{3}{*}{Optical flow}     & RAFT           &  & 0.6589                  & 0.7456                         & 0.9724                       & 0.9099                        & 0.9763                        &                      & 0.7528                  & 0.8093                         & 0.9819                       & 0.9190                        & 0.9898                        \\
                                        & RAFT (F)   &  & 0.7133                  & 0.8054                         & 0.9818                       & 0.9449                        & 0.9852                        &                      & 0.7758                  & 0.8424                         & 0.9860                       & 0.9425                        & 0.9922                        \\
                                        & RAFT (C) &  & 0.7028                  & 0.7944                         & 0.9723                       & 0.9411                        & 0.9690                        &                      & 0.7718                  & 0.8356                         & 0.9833                       & 0.9375                        & 0.9916                        \\ \cline{1-2} \cline{4-8} \cline{10-14} 
Inclusion matching                      & Ours                    &  & \textbf{0.8266}                  & \textbf{0.8726}                         & \textbf{0.9905}                       & \textbf{0.9724}                        & \textbf{0.9948}                        &                      & \textbf{0.8593}                  & \textbf{0.8929}                       & \textbf{0.9900}                       & \textbf{0.9638}                        & \textbf{0.9984}                        \\ \hline
\end{tabular}
}
\vspace{-5mm}
\end{table*}

\noindent{\bf Loss function.}
After the propagation step, the similarity matrix \(\textbf{S} \in \mathbb{R}^{N_j \times K_i}\) is represented as follows:
\begin{equation}
\textbf{S}_{mn} = \frac{\exp(\mathbf{\hat x}_{im} \cdot \mathbf{\hat x}_{jn})}{\sum_{m=1}^{N_j} \exp(\mathbf{\hat x}_{im} \cdot \mathbf{\hat x}_{jn})}.
\end{equation}

This matrix can be interpreted as the color probability \(\hat {\mathbf{y}}_m \in \mathbb{R}^{K_i}\) for segment \(m\) in the target frame. Given that each frame has a one-hot ground truth index label \(\mathbf{y}_m\), the loss function is derived from the cross-entropy loss:
\begin{equation}
L_{ce} = -\sum_{m=1}^{K_i} \mathbf{y}_m \log(\hat{\mathbf{y}}_m).
\end{equation}

\begin{table*}[]
\vspace{-3mm}
\caption{
Quantitative comparison of different datasets, different network structures, and different methods on test data. `IM' signifies the inclusion matching pipeline, which incorporates adjacent index label merging and segment-label matching strategies. Since the inclusion matching pipeline and coarse warping module with RAFT can also be applied to other networks, we also adopt them to AnT for comparison to illustrate the generalization of these methods. 
} \label{tab:ablation}
\vspace{-3mm}
\resizebox{1.0\textwidth}{!}{
\begin{tabular}{llcccccccccccc}
\hline
                         &                          &  & \multicolumn{5}{c}{3D rendered test set}                                                                                                                  & \multicolumn{1}{l}{} & \multicolumn{5}{c}{Hand-drawn test set}                                                                                                                  \\ \cline{4-8} \cline{10-14} 
\multirow{-2}{*}{Type}   & \multirow{-2}{*}{Method} &  & \multicolumn{1}{c}{Acc}       & \multicolumn{1}{l}{Acc-Thres} & \multicolumn{1}{l}{Pix-Acc}   & \multicolumn{1}{l}{Pix-F-Acc} & \multicolumn{1}{l}{Pix-B-MIoU} & \multicolumn{1}{l}{} & \multicolumn{1}{c}{Acc}       & \multicolumn{1}{l}{Acc-Thres} & \multicolumn{1}{l}{Pix-Acc}   & \multicolumn{1}{l}{Pix-F-Acc} & \multicolumn{1}{l}{Pix-B-MIoU} \\ \cline{1-8} \cline{10-14} 
Baseline                 & Ours                     &  & \textbf{0.8266}               & \textbf{0.8726}               & \textbf{0.9905}               & \textbf{0.9724}               & \textbf{0.9948}                & \textbf{}            & \textbf{0.8593}               & \textbf{0.8929}               & \textbf{0.9900}               & \textbf{0.9638}               & \textbf{0.9984}                \\ \cline{1-2} \cline{4-8} \cline{10-14} 
Dataset                  & AnimeRun                 &  & 0.7359                        & 0.7958                        & 0.9693                        & 0.9201                        & 0.9765                         &                      & 0.8278                        & 0.8634                        & 0.9818                        & 0.9399                        & 0.9906                         \\ \cline{1-2} \cline{4-8} \cline{10-14} 
                         & w/o CLIP                  &  & 0.8180                        & 0.8653                        & 0.9873                        & 0.9685                        & 0.9910                         &                      & 0.8489                        & 0.8843                        & 0.9878                        & 0.9570                        & 0.9925                         \\
                         & w/o RAFT                  &  & 0.8001                        & 0.8447                        & 0.9824                        & 0.9481                        & 0.9878      && 0.8446                        & 0.8750                        & 0.9804                        & 0.9404                        & 0.9851                                                                      \\
\multirow{-3}{*}{Model}  & w/o DConv                 &  &  0.8247 &  0.8685 &  0.9894 &  0.9709 & 0.9928                         &                      &  0.8525 &  0.8833 &  0.9875 &  0.9549 & 0.9937                         \\ \cline{1-2} \cline{4-8} \cline{10-14} 
                         & Ours w/o IM                    &  & 0.7637                        & 0.8352                        & 0.9721                        & 0.9573                        & 0.9688                         &                      & 0.8348                        & 0.8763                        & 0.9777                        & 0.9495                        & 0.9822                         \\
                         & AnT                      &  & 0.7450                        & 0.7930                        & 0.9708                        & 0.9361                        & 0.9758                         &                      & 0.8162                        & 0.8497                        & 0.9817                        & 0.9374                        & 0.9905                         \\
                         & AnT with IM                 &  & 0.8063                        & 0.8447                        & 0.9837                        & 0.9551                        & 0.9887                         &                      & 0.8388                        & 0.8698                        & 0.9839                        & 0.9388                        & 0.9943                         \\

\multirow{-4}{*}{Method} & AnT with IM/RAFT             &  & 0.8177    & 0.8587    & 0.9879    & 0.9681    & 0.9932     &  & 0.8483   & 0.8828   & 0.9886    & 0.9598    & 0.9946     \\ \hline
\end{tabular}
}
\vspace{-6mm}
\end{table*}

\noindent{\bf Inference Pipeline.} 
%Color probability accumulation+ Color redistribution
Since real animation's colors have an obvious domain gap with our randomly generated colors, we adopt a color redistribution strategy to reassign randomly generated colors to replace the original colors during the warping and matching pipeline. This strategy can avoid the mismatching problem when several predefined colors are too closed. Finally, the estimated color for each line-enclosed segment $\hat{\textbf{c}}_j \in \mathbb{R}^{N_j \times M}$ can be derived by:
\begin{equation}
\hat{\textbf{c}}_j = \hat{\textbf{S}} \textbf{c}_i,
\end{equation}
where $\hat{\textbf{S}} \in \mathbb{R}^{N_j \times N_i}$ is the similarity matrix for each line-enclosed segment in reference and target frames.  In this paper, we set the match threshold as 0.2 to filter out all the segments which are unmatchable.

%-------------------------Section Fig------------------------------

\section{Experiments}

\subsection{Implementation Details}
During the training stage, our model undergoes 300,000 iterations with a batch size of 2, utilizing NVIDIA GeForce RTX 3090 GPUs. We employ the Adam optimizer with a learning rate of $10^{-4}$ and no weight decay. Throughout the training process, we keep the weights of the pretrained optical flow estimation module and the CLIP image encoder fixed.
In the feature extraction module, the offset estimation module is structured as a lightweight U-Net with three down-sampling layers and a bottleneck featuring 128 features. Simultaneously, the feature extraction network is implemented as a U-Net with four down-sampling layers and a bottleneck containing 512 features. The channel number $C$ for our extracted feature is set to 128.
Within the multiplex transformer, the layer number $N$ is set to nine, and the number of heads is configured as four. The data augmentation pipeline is shown in the supplementary material.

\subsection{Comparisons with Previous Methods}
\begin{figure}[t]
  \centering
   \includegraphics[width=1.0\linewidth]{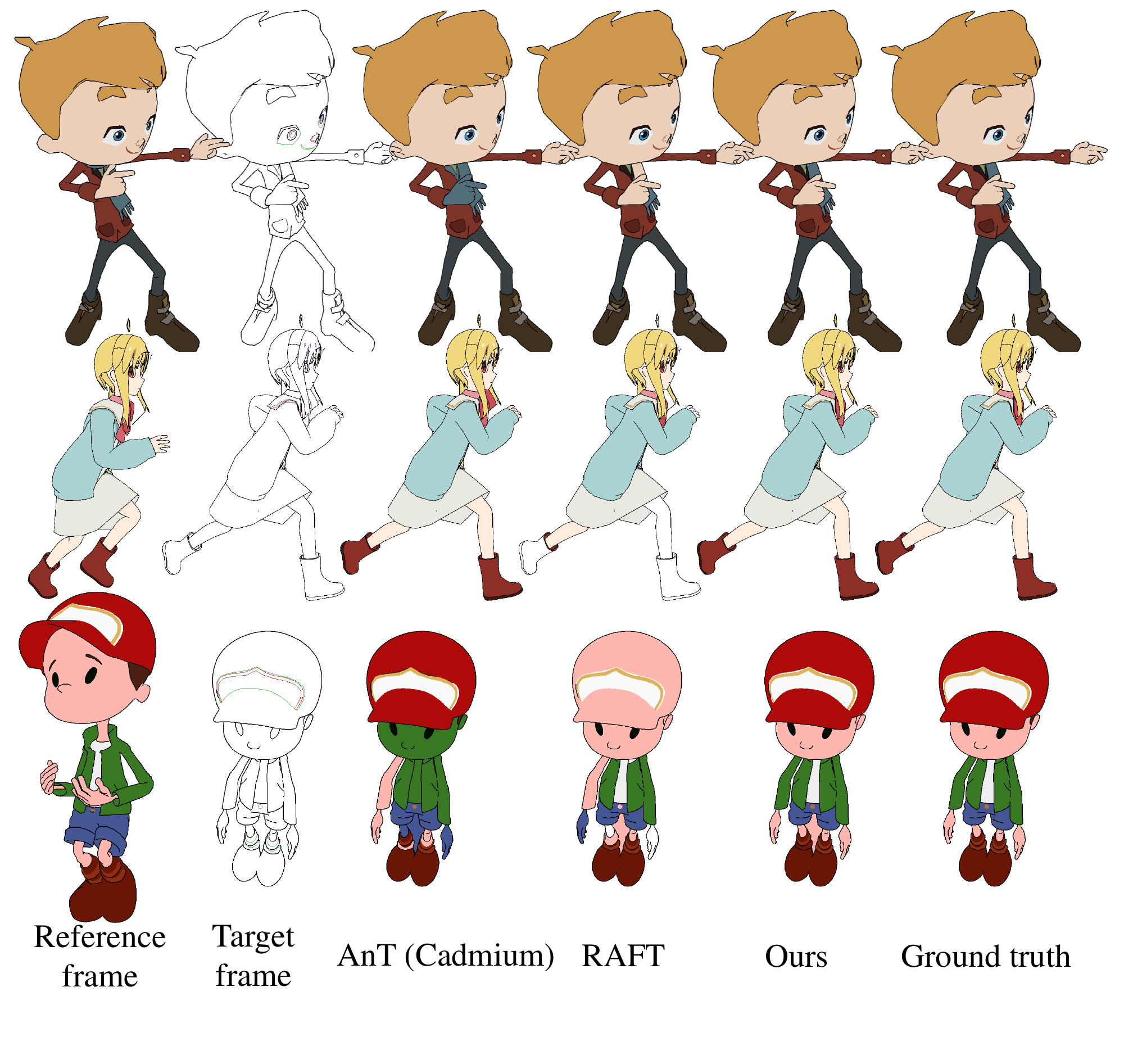}
   \vspace{-13mm}
   \caption{Visual comparison of different methods on our 3D rendered test set. Our proposed method produces satisfactory results in the presence of occlusion, significant motion, and substantial changes in viewing angles.}
   \vspace{-8mm}
   \label{fig:synthetic_anime}
\end{figure}

Since there is no existing paint bucket colorization method, we mainly compare our method with reference-based line art colorization~\cite{controlnet}, segment matching methods~\cite{AnT,siyao2022animerun}, and optical flow estimation methods~\cite{teed2020raft}.
The quantitative comparison is shown in Table~\ref{tab:baseline}. The visual comparison of real hand-drawn animation and rendered 3D test sets are shown in Fig.~\ref{fig:real_anime} and  Fig.~\ref{fig:synthetic_anime}.

\noindent{\bf Reference-based Line Art Colorization.}
The saturated luminance of the blank regions within line art impedes the application of reference-based methods for natural gray-scale image colorization.
Thus, we choose ControlNet~\cite{controlnet} as our reference-based line art colorization method.
We use Stable Diffusion v1.5~\cite{stable-diffusion} as our base model and employ the ControlNet's line art checkpoint to encode the line art condition. Then, we adopt the IP-Adapter~\cite{ye2023ip-adapter} to encode the reference image. We set the number of steps to 50 and use the text prompt `A character in the white background'.
We calculate the average RGB value per segment and match the average color with the nearest color in the reference image for evaluation.

\noindent{\bf Optical Flow Estimation.}
Since optical flow methods cannot produce segment-wise estimation without post-processing, we warp the color and calculate the colors that appear most frequently in each segment as the baseline of optical estimation method.
Since finetuning RAFT on AnimeRun can promote the optical flow's performance effectively, we use the finetuned RAFT in the following sessions.
%for all experiments

\noindent{\bf Segment Matching.} We also compare our method with the Cadmium application, the official implementation of AnT~\cite{AnT}, and AnimenRun's reimplemented AnT~\cite{siyao2022animerun}.
The visual comparison shown in Fig.~\ref{fig:real_anime} and  Fig.~\ref{fig:synthetic_anime} demonstrate the robust performance of our proposed method, particularly in scenarios where strict segment correspondence is absent, and challenges with large motion and complex deformations.
More visual results are presented in the suppl. material.

\subsection{Ablation Study}

\noindent{\bf Dataset.} To demonstrate the effectiveness of our proposed dataset, we also train the proposed method on the AnimeRun. In the training process, we adopt the inclusion matching strategy by merging the segments in the reference frames. The quantitative results reported in Table~\ref{tab:ablation} suggest the positive influence of our proposed dataset.

\noindent{\bf Model.} We conduct an ablation study on our model design to examine the necessity of each structure. In our experiments, we explore alternative configurations by replacing the deformable convolution and offset estimation module with the original convolution, eliminating the coarse warping module, and omitting the CLIP image encoder for ablation. The results presented in Table~\ref{tab:ablation} underscore the significance of each module, highlighting the crucial role of the coarse warping module. 

\noindent{\bf Method.} The results in Table~\ref{tab:ablation} further highlight the efficacy of our inclusion matching pipeline. Its application to AnT~\cite{AnT} yields a substantial improvement in network performance, suggesting the broad applicability and generalization of our inclusion matching pipeline.
Ablation study visualizations are available in the supplementary material.

\section{Conclusion}
We have introduced a new approach for line art paint bucket colorization. Departing from traditional segment matching methods that struggle with one-to-one correspondence, we proposed a novel inclusion matching pipeline that estimates each segment's inclusion relationship rather than the direct correspondence.
To facilitate the learning of this inclusion relationship, we presented a dedicated dataset, \pbc, and a two-stage colorization pipeline. Our pipeline incorporates a coarse color warping module and an inclusion matching module. The experimental results demonstrate the superior performance of our method in handling occlusion and accommodating large movements within challenging scenarios.

\noindent{\bf Acknowledgement.} This study is supported under the RIE2020 Industry Alignment Fund Industry Collaboration Projects (IAF-ICP) Funding Initiative, as well as cash and in-kind contribution from the industry partner(s).

% \fi

{
    \small
    \bibliographystyle{ieeenat_fullname}
    \bibliography{main}

\begin{thebibliography}{45}
\providecommand{\natexlab}[1]{#1}
\providecommand{\url}[1]{\texttt{#1}}
\expandafter\ifx\csname urlstyle\endcsname\relax
  \providecommand{\doi}[1]{doi: #1}\else
  \providecommand{\doi}{doi: \begingroup \urlstyle{rm}\Url}\fi

\bibitem[apl()]{aplaybox}
Aplaybox.
\newblock \url{https://www.aplaybox.com/}.

\bibitem[cad()]{cadmium}
Cadmium.
\newblock \url{https://cadmium.app/}.

\bibitem[mix()]{mixamo}
Mixamo.
\newblock \url{https://www.mixamo.com/}.

\bibitem[Akita et~al.(2023)Akita, Morimoto, and Tsuruno]{akita2023hand}
Kenta Akita, Yuki Morimoto, and Reiji Tsuruno.
\newblock Hand-drawn anime line drawing colorization of faces with texture details.
\newblock \emph{Computer Animation and Virtual Worlds}, 2023.

\bibitem[Butler et~al.(2012)Butler, Wulff, Stanley, and Black]{sintel}
D.~J. Butler, J. Wulff, G.~B. Stanley, and M.~J. Black.
\newblock A naturalistic open source movie for optical flow evaluation.
\newblock In \emph{ECCV}, 2012.

\bibitem[Cao et~al.(2021)Cao, Mo, and Gao]{cao2021line}
Ruizhi Cao, Haoran Mo, and Chengying Gao.
\newblock Line art colorization based on explicit region segmentation.
\newblock In \emph{Computer Graphics Forum}, 2021.

\bibitem[Cao et~al.(2023)Cao, Meng, Mok, Liu, Lee, and Li]{cao2023animediffusion}
Yu Cao, Xiangqiao Meng, PY Mok, Xueting Liu, Tong-Yee Lee, and Ping Li.
\newblock Animediffusion: Anime face line drawing colorization via diffusion models.
\newblock \emph{arXiv preprint arXiv:2303.11137}, 2023.

\bibitem[Casey et~al.(2021)Casey, P{\'e}rez, and Li]{AnT}
Evan Casey, V{\'\i}ctor P{\'e}rez, and Zhuoru Li.
\newblock The animation transformer: Visual correspondence via segment matching.
\newblock In \emph{ICCV}, 2021.

\bibitem[Chen et~al.(2020)Chen, Zhang, Gao, He, Xia, Shi, and Zhang]{chen2020active}
Shu-Yu Chen, Jia-Qi Zhang, Lin Gao, Yue He, Shihong Xia, Min Shi, and Fang-Lue Zhang.
\newblock Active colorization for cartoon line drawings.
\newblock \emph{TVCG}, 28\penalty0 (2), 2020.

\bibitem[Cherti et~al.(2023)Cherti, Beaumont, Wightman, Wortsman, Ilharco, Gordon, Schuhmann, Schmidt, and Jitsev]{cherti2023reproducible}
Mehdi Cherti, Romain Beaumont, Ross Wightman, Mitchell Wortsman, Gabriel Ilharco, Cade Gordon, Christoph Schuhmann, Ludwig Schmidt, and Jenia Jitsev.
\newblock Reproducible scaling laws for contrastive language-image learning.
\newblock In \emph{CVPR}, 2023.

\bibitem[Dai et~al.(2017)Dai, Qi, Xiong, Li, Zhang, Hu, and Wei]{dai2017deformable}
Jifeng Dai, Haozhi Qi, Yuwen Xiong, Yi Li, Guodong Zhang, Han Hu, and Yichen Wei.
\newblock Deformable convolutional networks.
\newblock In \emph{ICCV}, 2017.

\bibitem[Dang et~al.(2020)Dang, Do, Nguyen, Pham, Nguyen, Hoang, and Nguyen]{dang_correspondence_2020}
Trung D.~Q. Dang, Thien Do, Anh Nguyen, Van Pham, Quoc Nguyen, Bach Hoang, and Giao Nguyen.
\newblock Correspondence neural network for line art colorization.
\newblock In \emph{{SIGGRAPH}}, 2020.

\bibitem[Hu et~al.(2022)Hu, Shen, Wallis, Allen-Zhu, Li, Wang, Wang, and Chen]{lora}
Edward~J Hu, Yelong Shen, Phillip Wallis, Zeyuan Allen-Zhu, Yuanzhi Li, Shean Wang, Lu Wang, and Weizhu Chen.
\newblock Lo{RA}: Low-rank adaptation of large language models.
\newblock In \emph{ICLR}, 2022.

\bibitem[Hu(1962)]{hu1962visual}
Ming-Kuei Hu.
\newblock Visual pattern recognition by moment invariants.
\newblock \emph{IRE transactions on information theory}, 8\penalty0 (2), 1962.

\bibitem[Kim et~al.(2019)Kim, Jhoo, Park, and Yoo]{kim_tag2pix_2019}
Hyunsu Kim, Ho~Young Jhoo, Eunhyeok Park, and Sungjoo Yoo.
\newblock {Tag2Pix}: Line art colorization using text tag with {SECat} and changing loss.
\newblock In \emph{ICCV}, 2019.

\bibitem[Lee et~al.(2020)Lee, Kim, Lee, Kim, Chang, and Choo]{lee2020reference}
Junsoo Lee, Eungyeup Kim, Yunsung Lee, Dongjun Kim, Jaehyuk Chang, and Jaegul Choo.
\newblock Reference-based sketch image colorization using augmented-self reference and dense semantic correspondence.
\newblock In \emph{CVPR}, 2020.

\bibitem[Liu et~al.(2020)Liu, Wang, Wu, and Seah]{liu2020shape}
Shaolong Liu, Xingce Wang, Zhongke Wu, and Hock~Soon Seah.
\newblock Shape correspondence based on kendall shape space and rag for 2d animation.
\newblock \emph{The Visual Computer}, 36, 2020.

\bibitem[Liu et~al.(2023)Liu, Wang, Liu, Wu, and Seah]{liu_shape_2023}
Shaolong Liu, Xingce Wang, Xiangyuan Liu, Zhongke Wu, and Hock~Soon Seah.
\newblock Shape correspondence for cel animation based on a shape association graph and spectral matching.
\newblock \emph{Computational Visual Media}, 2023.

\bibitem[Maejima et~al.(2019)Maejima, Kubo, Funatomi, Yotsukura, Nakamura, and Mukaigawa]{maejima2019graph}
Akinobu Maejima, Hiroyuki Kubo, Takuya Funatomi, Tatsuo Yotsukura, Satoshi Nakamura, and Yasuhiro Mukaigawa.
\newblock Graph matching based anime colorization with multiple references.
\newblock In \emph{SIGGRAPH}. 2019.

\bibitem[Ohira(2010)]{ae_for_animation}
Koki Ohira.
\newblock \emph{After Effects For Animation}.
\newblock 2010.

\bibitem[publishing department(2010{\natexlab{a}})]{kyoto_colorization}
Kyoto~Animation publishing department.
\newblock \emph{Guide Kyoto Animation version finish, shooting}.
\newblock Kyoto Animation Co., Ltd, 2010{\natexlab{a}}.

\bibitem[publishing department(2010{\natexlab{b}})]{kyoto_draw}
Kyoto~Animation publishing department.
\newblock \emph{The Kyoto Animation Guide to Drawing}.
\newblock Kyoto Animation Co., Ltd, 2010{\natexlab{b}}.

\bibitem[Radford et~al.(2021)Radford, Kim, Hallacy, Ramesh, Goh, Agarwal, Sastry, Askell, Mishkin, Clark, et~al.]{clip}
Alec Radford, Jong~Wook Kim, Chris Hallacy, Aditya Ramesh, Gabriel Goh, Sandhini Agarwal, Girish Sastry, Amanda Askell, Pamela Mishkin, Jack Clark, et~al.
\newblock Learning transferable visual models from natural language supervision.
\newblock In \emph{ICML}, 2021.

\bibitem[Rombach et~al.(2022)Rombach, Blattmann, Lorenz, Esser, and Ommer]{stable-diffusion}
Robin Rombach, Andreas Blattmann, Dominik Lorenz, Patrick Esser, and Bj{\"o}rn Ommer.
\newblock High-resolution image synthesis with latent diffusion models.
\newblock In \emph{CVPR}, 2022.

\bibitem[Ronneberger et~al.(2015)Ronneberger, Fischer, and Brox]{unet}
Olaf Ronneberger, Philipp Fischer, and Thomas Brox.
\newblock U-net: Convolutional networks for biomedical image segmentation.
\newblock In \emph{MICCAI}. Springer, 2015.

\bibitem[Ruiz et~al.(2023)Ruiz, Li, Jampani, Pritch, Rubinstein, and Aberman]{dreambooth}
Nataniel Ruiz, Yuanzhen Li, Varun Jampani, Yael Pritch, Michael Rubinstein, and Kfir Aberman.
\newblock Dreambooth: Fine tuning text-to-image diffusion models for subject-driven generation.
\newblock In \emph{CVPR}, 2023.

\bibitem[Sarlin et~al.(2020)Sarlin, DeTone, Malisiewicz, and Rabinovich]{sarlin20superglue}
Paul-Edouard Sarlin, Daniel DeTone, Tomasz Malisiewicz, and Andrew Rabinovich.
\newblock {SuperGlue}: Learning feature matching with graph neural networks.
\newblock In \emph{CVPR}, 2020.

\bibitem[Shugrina et~al.(2019)Shugrina, Liang, Kar, Li, Singh, Singh, and Fidler]{shugrina2019creative}
Maria Shugrina, Ziheng Liang, Amlan Kar, Jiaman Li, Angad Singh, Karan Singh, and Sanja Fidler.
\newblock Creative flow+ dataset.
\newblock In \emph{CVPR}, 2019.

\bibitem[Siyao et~al.(2021)Siyao, Zhao, Yu, Sun, Metaxas, Loy, and Liu]{siyao_deep_2021}
Li Siyao, Shiyu Zhao, Weijiang Yu, Wenxiu Sun, Dimitris Metaxas, Chen~Change Loy, and Ziwei Liu.
\newblock Deep animation video interpolation in the wild.
\newblock In \emph{CVPR}, 2021.

\bibitem[Siyao et~al.(2022)Siyao, Li, Li, Dong, Liu, and Loy]{siyao2022animerun}
Li Siyao, Yuhang Li, Bo Li, Chao Dong, Ziwei Liu, and Chen~Change Loy.
\newblock Animerun: 2d animation visual correspondence from open source 3d movies.
\newblock \emph{NeurIPS}, 2022.

\bibitem[Siyao et~al.(2023)Siyao, Gu, Xiao, Ding, Liu, and Loy]{siyao2023inbetween}
Li Siyao, Tianpei Gu, Weiye Xiao, Henghui Ding, Ziwei Liu, and Chen~Change Loy.
\newblock Deep geometrized cartoon line inbetweening.
\newblock In \emph{ICCV}, 2023.

\bibitem[S{\`y}kora et~al.(2004)S{\`y}kora, Buri{\'a}nek, and {\v{Z}}{\'a}ra]{sykora2004unsupervised}
Daniel S{\`y}kora, Jan Buri{\'a}nek, and Ji{\v{r}}{\'\i} {\v{Z}}{\'a}ra.
\newblock Unsupervised colorization of black-and-white cartoons.
\newblock In \emph{International Symposium on Non-photorealistic Animation and Rendering}, 2004.

\bibitem[S{\`y}kora et~al.(2009{\natexlab{a}})S{\`y}kora, Dingliana, and Collins]{sykora2009lazybrush}
Daniel S{\`y}kora, John Dingliana, and Steven Collins.
\newblock Lazybrush: Flexible painting tool for hand-drawn cartoons.
\newblock In \emph{Computer Graphics Forum}, 2009{\natexlab{a}}.

\bibitem[S{\`y}kora et~al.(2009{\natexlab{b}})S{\`y}kora, Dingliana, and Collins]{sykora2009rigid}
Daniel S{\`y}kora, John Dingliana, and Steven Collins.
\newblock As-rigid-as-possible image registration for hand-drawn cartoon animations.
\newblock In \emph{International Symposium on Non-photorealistic Animation and Rendering}, 2009{\natexlab{b}}.

\bibitem[Teed and Deng(2020)]{teed2020raft}
Zachary Teed and Jia Deng.
\newblock Raft: Recurrent all-pairs field transforms for optical flow.
\newblock In \emph{ECCV}, 2020.

\bibitem[Vaswani et~al.(2017)Vaswani, Shazeer, Parmar, Uszkoreit, Jones, Gomez, Kaiser, and Polosukhin]{vaswani2017attention}
Ashish Vaswani, Noam Shazeer, Niki Parmar, Jakob Uszkoreit, Llion Jones, Aidan~N Gomez, {\L}ukasz Kaiser, and Illia Polosukhin.
\newblock Attention is all you need.
\newblock \emph{NeurIPS}, 30, 2017.

\bibitem[Wu et~al.(2023)Wu, Yan, Liu, Xu, and Zhang]{wu2023self}
Shukai Wu, Xiao Yan, Weiming Liu, Shuchang Xu, and Sanyuan Zhang.
\newblock Self-driven dual-path learning for reference-based line art colorization under limited data.
\newblock \emph{IEEE TCSVT}, 2023.

\bibitem[Ye et~al.(2023)Ye, Zhang, Liu, Han, and Yang]{ye2023ip-adapter}
Hu Ye, Jun Zhang, Sibo Liu, Xiao Han, and Wei Yang.
\newblock Ip-adapter: Text compatible image prompt adapter for text-to-image diffusion models.
\newblock \emph{arXiv preprint arxiv:2308.06721}, 2023.

\bibitem[Zhang et~al.(2012)Zhang, Huang, and Fu]{zhang_excol_2012}
Lei Zhang, Hua Huang, and Hongbo Fu.
\newblock {EXCOL}: An {EXtract}-and-{COmplete} layering approach to cartoon animation reusing.
\newblock \emph{TVCG}, 18\penalty0 (7), 2012.

\bibitem[Zhang et~al.(2018)Zhang, Li, Wong, Ji, and Liu]{xia-2018-invertible}
Lvmin Zhang, Chengze Li, Tien-Tsin Wong, Yi Ji, and Chunping Liu.
\newblock Two-stage sketch colorization.
\newblock \emph{ACM TOG}, 37\penalty0 (6), 2018.

\bibitem[Zhang et~al.(2021{\natexlab{a}})Zhang, Li, Simo-Serra, Ji, Wong, and Liu]{Filling2021zhang}
Lvmin Zhang, Chengze Li, Edgar Simo-Serra, Yi Ji, Tien-Tsin Wong, and Chunping Liu.
\newblock User-guided line art flat filling with split filling mechanism.
\newblock In \emph{CVPR}, 2021{\natexlab{a}}.

\bibitem[Zhang et~al.(2023)Zhang, Rao, and Agrawala]{controlnet}
Lvmin Zhang, Anyi Rao, and Maneesh Agrawala.
\newblock Adding conditional control to text-to-image diffusion models.
\newblock In \emph{ICCV}, 2023.

\bibitem[Zhang et~al.(2021{\natexlab{b}})Zhang, Wang, Wen, Li, and Liu]{zhang2021line}
Qian Zhang, Bo Wang, Wei Wen, Hai Li, and Junhui Liu.
\newblock Line art correlation matching feature transfer network for automatic animation colorization.
\newblock In \emph{WACV}, 2021{\natexlab{b}}.

\bibitem[Zhu et~al.(2016)Zhu, Liu, Wong, and Heng]{zhu-2016-toontrack}
Haichao Zhu, Xueting Liu, Tien-Tsin Wong, and Pheng-Ann Heng.
\newblock Globally optimal toon tracking.
\newblock \emph{ACM TOG}, 35\penalty0 (4), 2016.

\bibitem[Zou et~al.(2019)Zou, Mo, Gao, Du, and Fu]{zouSA2019sketchcolorization}
Changqing Zou, Haoran Mo, Chengying Gao, Ruofei Du, and Hongbo Fu.
\newblock Language-based colorization of scene sketches.
\newblock \emph{ACM TOG}, 38\penalty0 (6), 2019.

\end{thebibliography}
}

\clearpage

\twocolumn[{%
\renewcommand\twocolumn[1][]{#1}%
\setcounter{page}{1}
\maketitlesupplementary
\begin{center}
    \centering
    \captionsetup{type=figure}
    \vspace{-2mm}
    \includegraphics[width=1.0\linewidth]{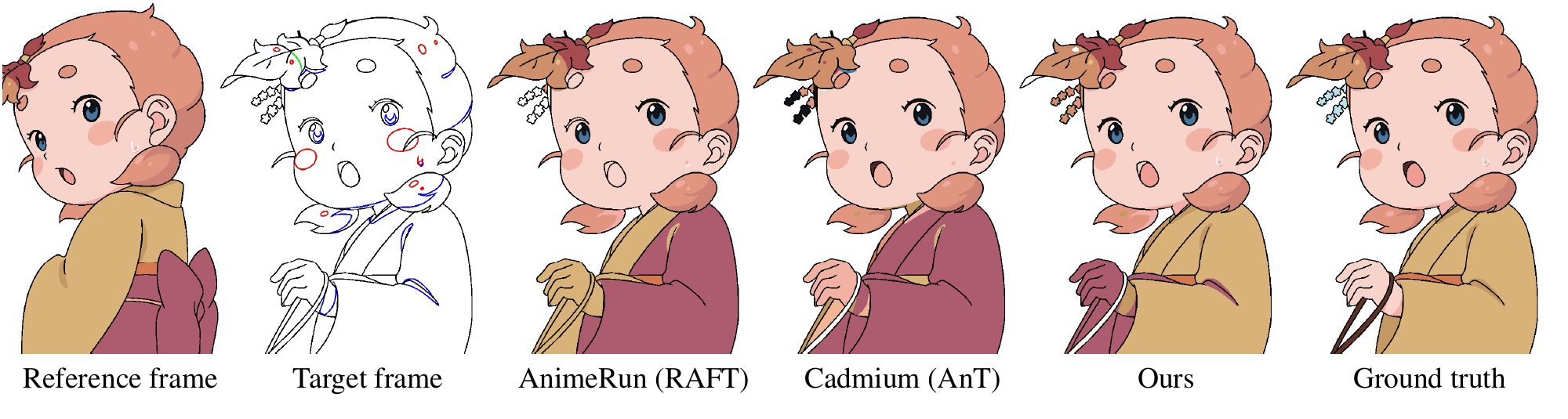}
   \vspace{-10mm}
   \caption{Visual comparison of different methods' performances on challenging scenes.}
   \label{fig:failure}
\end{center}%
}]

\renewcommand\thesection{\Alph{section}}
\setcounter{section}{0}

\section{Data Augmentation}
Different from optical flow annotations provided in AnimeRun~\cite{siyao2022animerun}, index labels in our dataset can produce direct inclusion matching correspondences even for non-adjacent frames.
Notably, even when the reference frame and target frame are the same frame, the inclusion relationship still may not be apparent to the network. During the training process, we set the inter-frame intervals as 0, 1, and 2, respectively, as a form of data augmentation.
Additionally, we employ separate random translation in $U(-400,400)$, random rotation in $U(-\pi/{9},\pi/9)$, and random resize in $U(1/2,1/3)$ for paired frames to emulate typical movements in animation. 
%Notably, distinct augmentations are applied to the target and reference frames. 
In the resizing augmentation, a min pooling strategy is implemented to prevent the removal of some pixels in black lines through nearest or bilinear interpolation.
After these augmentation steps, all training data undergo cropping to achieve a consistent size of $640 \times 640$.

\section{Limitation}
%Complex occlusion relationships
%Exaggerated motion 
%shadow and highlight always change shape
While our inclusion matching pipeline addresses many-to-many matching scenarios, it may still encounter challenges in complex occlusion situations and when faced with exaggerated motion or distortion, potentially leading to incorrect matching. Existing reference-based pipelines necessitate the presence of all segments from the target frame in the reference frame, making it difficult to accurately colorize new segments in the scenes like opening eyes or turning around, as the hand illustrated in Fig.~\ref{fig:failure}. Despite these challenges, our method still shows superior performance compared to existing approaches in such demanding scenarios.

\begin{figure}[t]
  \centering
    \includegraphics[width=1.0\linewidth]{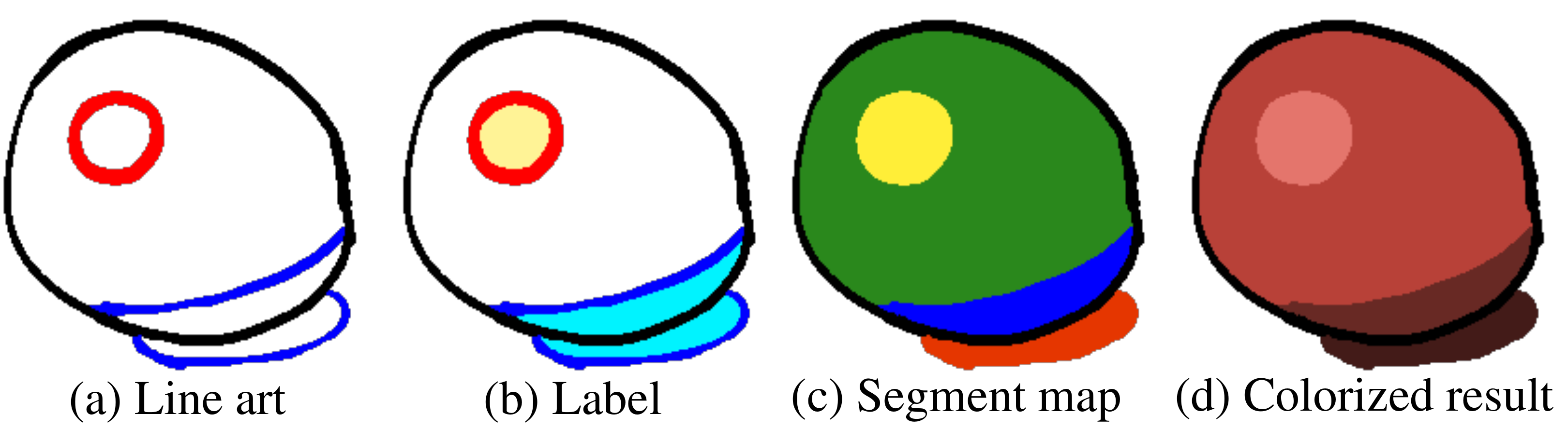}
   \vspace{-4mm}
   \caption{In the animation industry, color lines need to be colorized as the colors of the adjacent highlight and shadow regions as (d). Besides, the animators will label the highlight and shadow with yellow and blue as shown in (b) of each segment to avoid confusion. Thus, in the evaluation, we merge the color lines with the adjacent highlight and shadow as the segment map (c) to avoid estimating these color lines' colors.}
   \vspace{-6mm}
   \label{fig:highlight_label}
\end{figure}

\section{Color Line Colorization} 
\if 0
How to color the color line.
three ways:

(1) Merge the line with the adjacent segment with most small area.

(2) expand the region to colorize the line.

(3) Use network to classify the color of each line. 
\fi

In the animation industry, color lines (such as the blue and red lines in Fig.~\ref{fig:highlight_label}) are often used to mark shadows and highlights. 
% These lines also need to be colorized.
In general, these color lines also need to be colorized as the adjacent shadow or highlight's colors~\cite{kyoto_colorization}. This routine ensures a seamless integration of the color lines with the surrounding visual elements.
Furthermore, the animators are required to label the highlight and shadow regions with yellow and blue as shown in Fig.~\ref{fig:highlight_label}(b) to prevent any potential confusion among downstream digital painters in animation production standards~\cite{kyoto_draw}.
Thus, the segments of color lines can be merged into the adjacent highlight or shadow segment for inclusion matching.
This approach eliminates the necessity to independently predict the color of each color line.
For evaluation in the Table~3 and Table~4 of the main paper, the annotations of our segment map which we use to calculate the segment accuracy are derived from calculating the connected component of each color in the colorized ground truth.
This design prevents our benchmark from penalizing methods that do not specifically cater to color line colorization.

\section{More Analysis}

\begin{figure*}[t]
  \centering
   \includegraphics[width=1.0\linewidth]{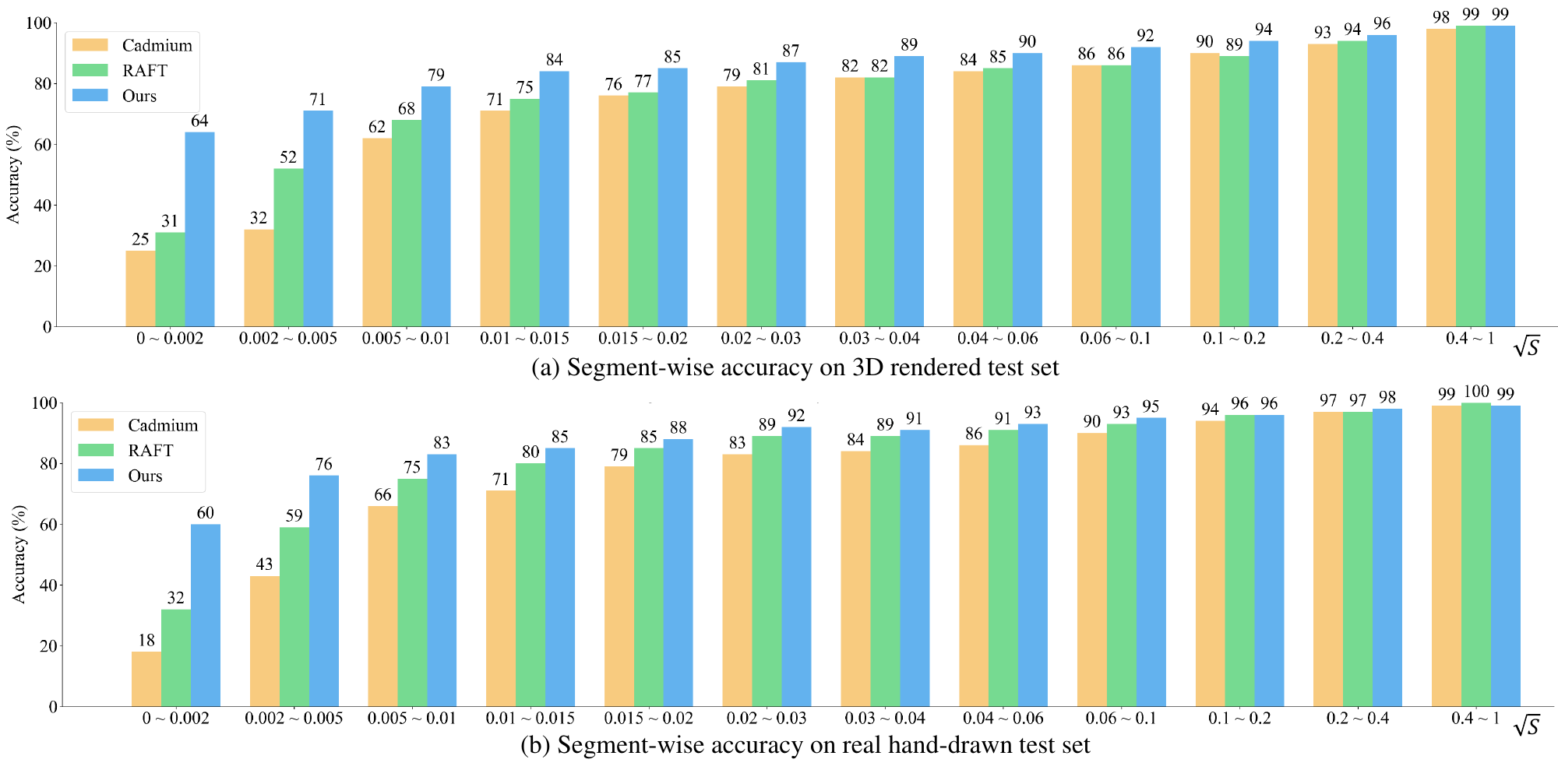}
   \vspace{-8mm}
   \caption{Accuracy comparisons for segments of different sizes. We calculate the segment-wise accuracy in different bins, each partitioned according to the proportional area of the respective segment, denoted as $S$. Since the majority of segments have small areas, we partition the intervals using $\sqrt{S}$ and ensure a roughly equal number of segments in each bin. Same as the experiment in the main paper, the RAFT~\cite{teed2020raft} is finetuned on the AnimeRun~\cite{siyao2022animerun} dataset, and the official implementation of AnT~\cite{AnT} is realized through the Cadmium application~\cite{cadmium}.}
   \label{fig:seg_accu_label}
\end{figure*}

\begin{figure*}[t]
  \centering
   \includegraphics[width=1.0\linewidth]{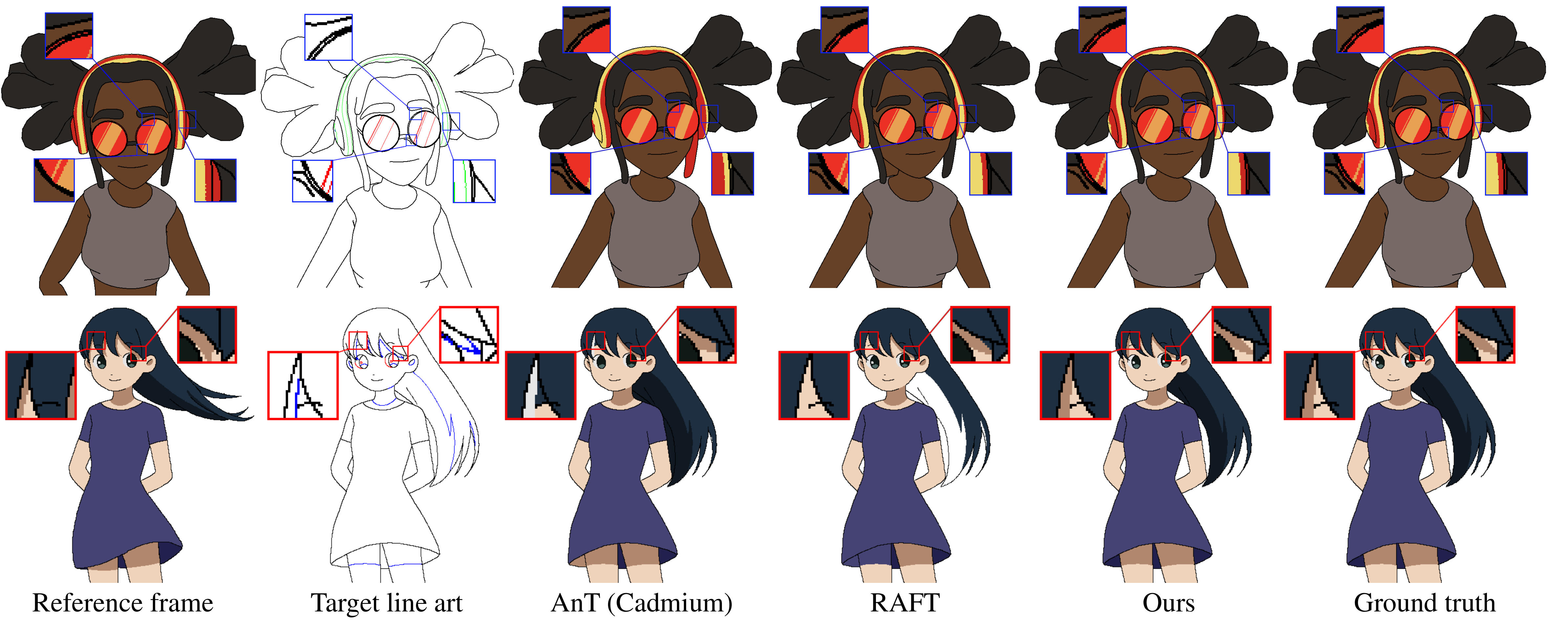}
   \vspace{-8mm}
   \caption{Visual comparisons of tiny segment colorization across different methods. Even in scenarios with no corresponding segment in the reference frame, our approach still excels in colorizing tiny segments accurately.}
   \vspace{-3mm}
   \label{fig:tiny_seg}
\end{figure*}

\noindent{\bf Colorization for Tiny Segments.}
% table + figure
% inference self_prop
Colorizing tiny segments poses a significant challenge in paint bucket colorization, consuming a considerable amount of time for digital painters. Despite the existence of methods based on segment matching~\cite{AnT} and optical flow, accurate colorization of tiny segments remains challenging.
In Fig.~\ref{fig:seg_accu_label}, we evaluate the accuracy of segment colorization across various sizes. Our proposed method outperforms current optical-flow-based approaches~\cite{teed2020raft,siyao2022animerun} and AnT~\cite{AnT}, particularly excelling in tiny segments.
Even when dealing with segments as small as a few pixels, our inclusion matching training pipeline demonstrates its effectiveness in guiding the network to produce accurate and reliable results. The visual comparisons in Fig.~\ref{fig:tiny_seg} highlight the superior performance of our method in handling these challenging tiny segments.

\begin{figure*}[t]
  \centering
   \includegraphics[width=1.0\linewidth]{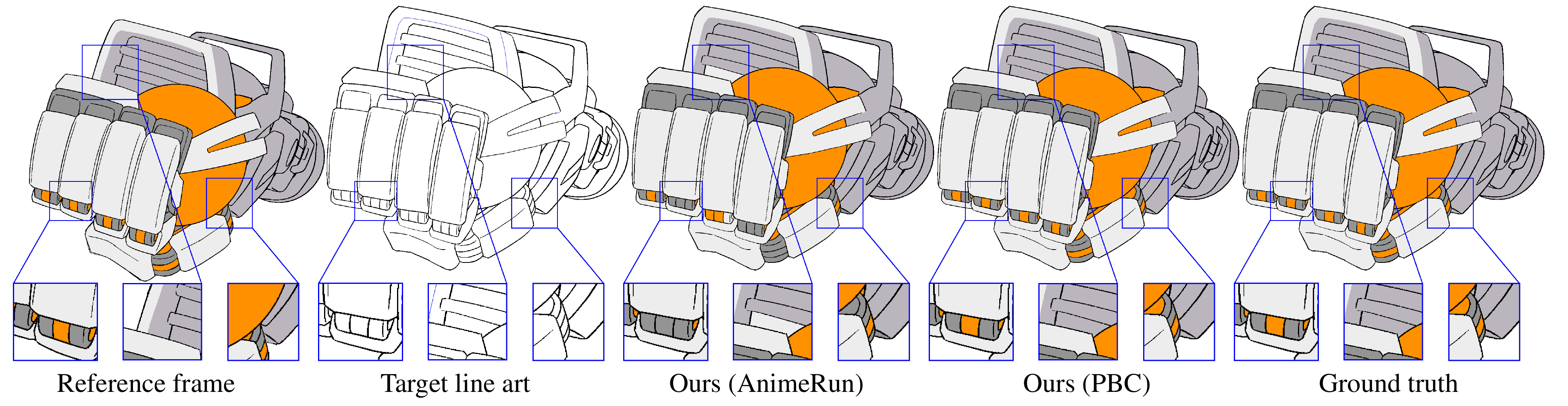}
   \vspace{-6mm}
   \caption{Visual comparison on training network with different datasets. `PBC' represents our proposed dataset \pbc. }
   \vspace{-3mm}
   \label{fig:ablation_dataset}
\end{figure*}

\begin{figure*}[t]
  \centering
   \includegraphics[width=1.0\linewidth]{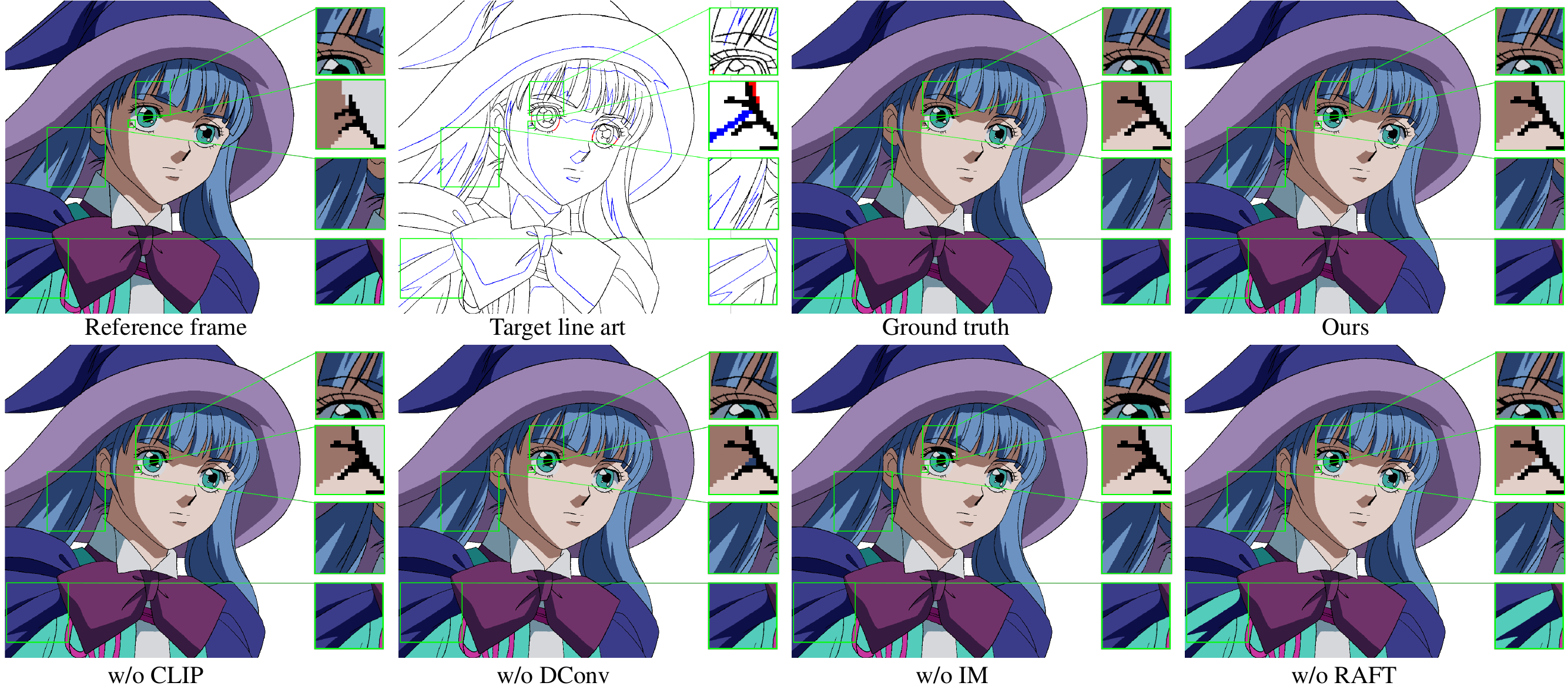}
   \vspace{-7mm}
   \caption{Ablation study for different modules of our method. Zoom in for better visualization.}
   \label{fig:ablation_study}
\end{figure*}

\begin{figure*}[t]
  \centering
   \includegraphics[width=1.0\linewidth]{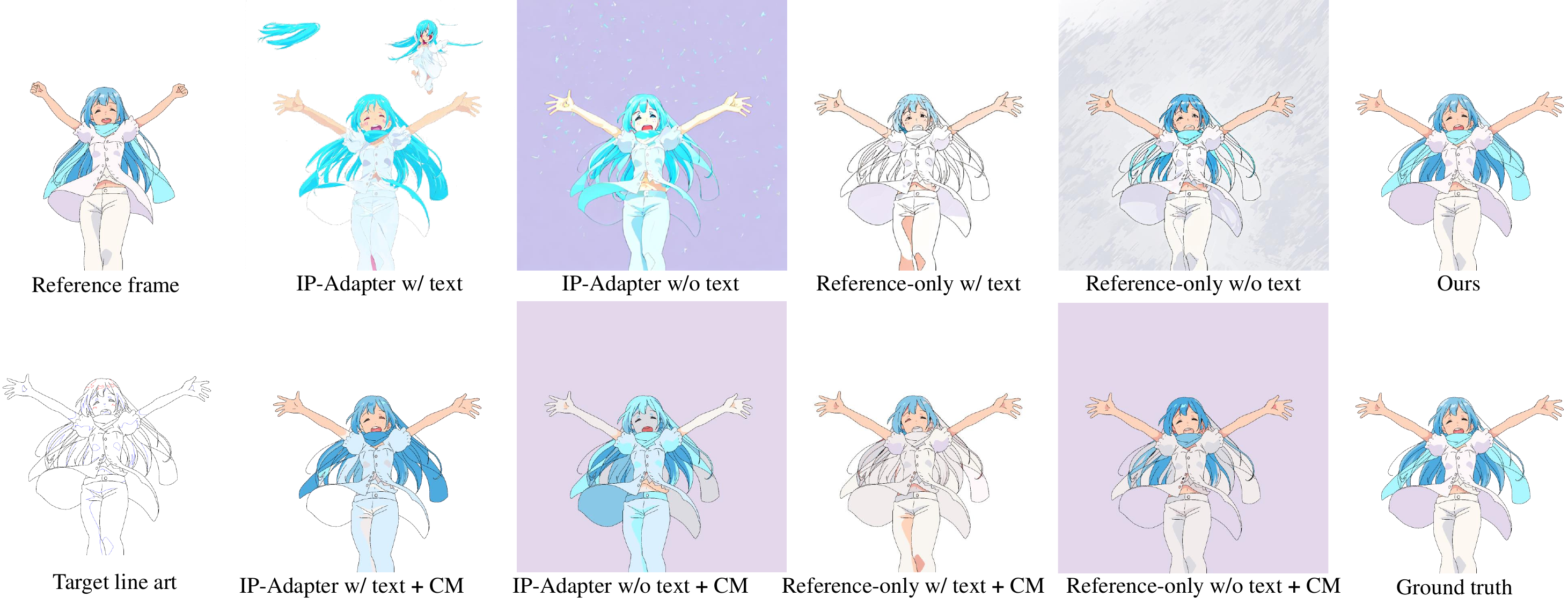}
   \vspace{-6mm}
   \caption{Visual comparisons of our method and different reference-based pixel-wise line art colorization methods. `w/ \& w/o text' means using or not using the text prompt `A character in the white background'. In the second row, we employ `CM' (color mapping), a technique that maps the average color in each segment to the closest color in the reference frame. Compared with these pixel-wise colorization methods, our method can yield more reliable results by avoiding filling similar colors in segments.}
   \vspace{-2mm}
   \label{fig:controlnet}
\end{figure*}

\begin{figure*}[t]
  \centering
   \includegraphics[width=0.85\linewidth]{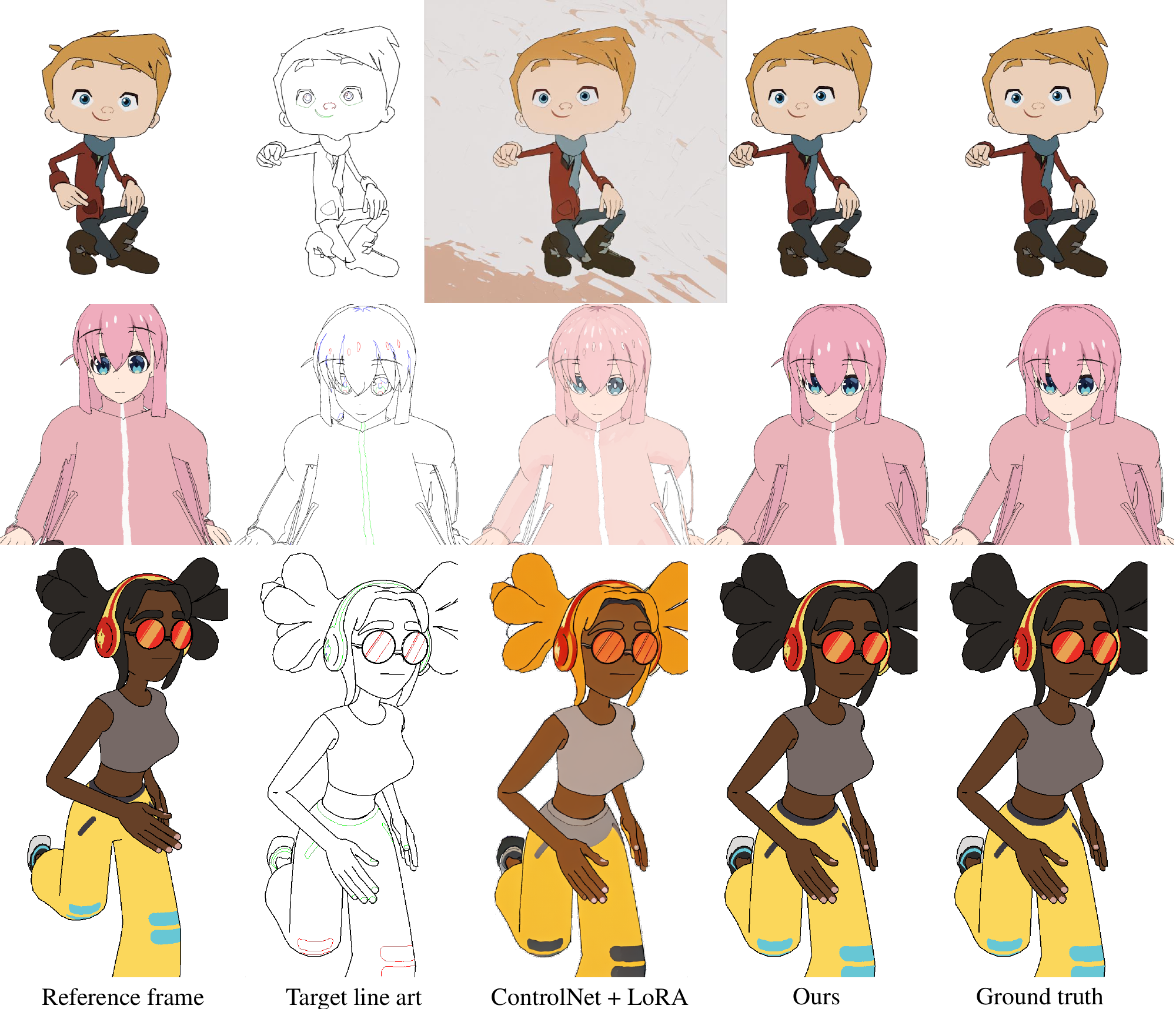}
   \vspace{-2mm}
   \caption{Visual comparisons of our method and finetuning-based pixel-wise line art colorization method. We finetune the LoRA~\cite{lora} for each character using DreamBooth's pipeline and apply ControlNet~\cite{controlnet} to control the contour of the colorized result.}
   \vspace{-6mm}
   \label{fig:lora}
\end{figure*}

%\noindent{\bf Colorization for large motion.}
\noindent{\bf Visualization of Ablation Study.} In Fig.~\ref{fig:ablation_dataset}, we train our baseline model with AnimeRun and our proposed \pbc~dataset. Compared with the baseline model trained on AnimeRun, our method can achieve more robust visualization results. Although our dataset only consists of anime characters, Fig.~\ref{fig:ablation_dataset} demonstrates that our method can also generalize well on different scenes. Also, we conduct an ablation study on deformable convolution (w/o DConv), coarse warping module (w/o RAFT), inclusion matching pipeline (w/o IM), and CLIP image encoder (w/o CLIP). Fig.~\ref{fig:ablation_study} particularly highlights the crucial role of the coarse warping module and inclusion matching. 
The warped color feature introduces a leakage of grouping information from the reference frame, facilitating the feature extraction network in learning inclusion relationships more effectively compared to relying solely on urging the multiplex transformer to acquire this representation.
Thus, lacking coarse warping or inclusion matching will all lead to mismatching while there is no one-to-one correspondence as shown in the eyelash part of Fig.~\ref{fig:ablation_study}.

\noindent{\bf Reference-based Pixel-wise Approaches.}  Latent diffusion~\cite{stable-diffusion} can support different types of ways to encode the condition of the reference image. In our experiments, we compare the performance of IP-Adapter~\cite{ye2023ip-adapter} and ControlNet's reference-only preprocessor~\cite{controlnet} as reference frame encoder in Fig.~\ref{fig:controlnet}. Also, we test using or not using text prompts in this figure. The target line art is encoded using ControlNet's line art model.
Meanwhile, finetuning-based methods such as LoRA~\cite{lora} also serve as a potential solution for reference-based colorization.
We also finetune the additional LoRA~\cite{lora} for each character based on 9 frames of each character's T-pose and 1 reference frame following DreamBooth~\cite{dreambooth}. 
Then, we use ControlNet's line art model and finetuned LoRA to colorize the target line as shown in Fig.~\ref{fig:lora} with text prompt `A [V] character'.
The visualization results demonstrate that these ControlNet-based methods cannot restore accurate colors even with the color mapping strategy.
Besides, the limited resolution in latent diffusion's denoising process results in the fact that the colors of character details such as eyes are difficult to maintain.
Compared with these methods, our method can achieve more accurate results and can even function well on pixel-level details.


\end{document}